\documentclass{article} 
\usepackage{iclr2027_conference,times}

\usepackage{basic_commands}

\definecolor{myred}{HTML}{F54254}
\definecolor{myorange}{HTML}{FFB135}
\definecolor{mygreen}{HTML}{10BD35}
\definecolor{myblue}{HTML}{598BE7}
\definecolor{mypurple}{HTML}{9A1C6B}
\definecolor{plgray}{HTML}{999999}
\definecolor{col}{HTML}{598BE7}
\definecolor{col2}{HTML}{F54254}
\definecolor{col3}{RGB}{66, 244, 133}
\definecolor{promptblue}{HTML}{2457A6}

\definecolor{soheilblue}{HTML}{123A6B}
\newif\ifsoheilcmt\soheilcmttrue

\usepackage{hyperref}
\usepackage{url}
\usepackage{booktabs}
\usepackage{amsfonts}
\usepackage{nicefrac}
\usepackage{microtype}
\usepackage{xcolor}
\usepackage{dsfont}
\usepackage{caption}
\usepackage{subcaption}
\usepackage{graphicx}
\usepackage{wrapfig}
\usepackage{amsthm}
\usepackage{cleveref}
\usepackage{enumitem}
\usepackage{tcolorbox}
\usepackage{tikz}
\usepackage{multirow}
\usepackage{makecell}
\usepackage{tablefootnote}
\usepackage{mathtools}
\usepackage{siunitx}
\usepackage{makecell}
\usepackage[para,online,flushleft]{threeparttable}
\usepackage{soul}
\usepackage{algorithm}
\usepackage{algorithmic}
\usetikzlibrary{arrows.meta}
\usepackage{adjustbox}
\usepackage{colortbl}
\usepackage{wrapfig}
\usepackage{needspace}
\usepackage{float}
\usepackage{xcolor,colortbl}
\usepackage{array}
\usepackage{longtable}
\usepackage{xcolor}
\usepackage{colortbl}
\usepackage{amsmath}
\usepackage{arydshln}
\newcommand{\ourscell}[1]{\cellcolor[HTML]{EFEFEF}#1}

\setlist[itemize]{itemsep=1pt, leftmargin=20pt}
\hypersetup{urlcolor=myblue,citecolor=myblue,linkcolor=myblue}

\usepackage{listings}
\lstdefinestyle{pythonstyle}{
    language=Python,           
    basicstyle=\ttfamily\small,
    keywordstyle=\color{black}, 
    stringstyle=\color{orange!60!black},   
    commentstyle=\color{green!50!black}, 
    numbers=left,              
    numberstyle=\tiny\color{gray}, 
    breaklines=true,           
    frame=single,              
    rulecolor=\color{black},   
    showstringspaces=false,    
    tabsize=4,                 
    captionpos=b,              
    breakatwhitespace=false,   
}

\lstdefinestyle{pythonstyle2}{
  language=Python,
  basicstyle=\ttfamily\tiny,
  keywordstyle=\color{blue}\bfseries,
  stringstyle=\color{orange!60!black},
  commentstyle=\color{green!50!black},
  numbers=none,
  breaklines=true,
  frame=none,
  showstringspaces=false,
  columns=fullflexible,
  xleftmargin = -12pt,
  xrightmargin = -12pt,
  framesep=0pt,
  framexleftmargin=-10pt,
  framexrightmargin=-10pt,
  aboveskip = -5pt,
  belowskip = -5pt,
  lineskip = -3pt,
  escapeinside={(@}{@)},
}

\lstdefinestyle{pythonstyle3}{
  backgroundcolor=\color{col!5},
  basicstyle=\ttfamily\footnotesize,
  keywordstyle=\color{black},
  stringstyle=\color{orange!60!black},
  commentstyle=\color{green!50!black},
  numbers=none,
  breaklines=true,
  frame=none,
  showstringspaces=false,
  columns=fullflexible,
  keepspaces=true,                 
  showspaces=false,
  framesep=0pt,
  aboveskip = 0pt,
  belowskip = 0pt,
  escapeinside={(@}{@)},
}

\lstdefinestyle{pythonstyle4}{
  backgroundcolor=\color{col!5},
  language=Python,
  basicstyle=\ttfamily\footnotesize,
  keywordstyle=\color{blue}\bfseries,
  stringstyle=\color{orange!60!black},
  commentstyle=\color{green!50!black},
  identifierstyle=\color{black},
  numbers=none,
  breaklines=true,
  frame=none,
  showstringspaces=false,
  columns=fullflexible,
  keepspaces=true,                 
  showspaces=false,
  framesep=0pt,
  aboveskip = 0pt,
  belowskip = 0pt,
  escapeinside={(@}{@)},
}

\lstdefinestyle{promptstyle}{
  backgroundcolor=\color{col!5},
  basicstyle=\ttfamily\scriptsize,
  numbers=none,
  breaklines=true,
  breakatwhitespace=true,
  frame=none,
  showstringspaces=false,
  columns=fullflexible,
  keepspaces=true,
  showspaces=false,
  framesep=0pt,
  aboveskip=0pt,
  belowskip=0pt,
  escapeinside={(@}{@)},
}

\title{Bias Control and Injection for Multi-Agent Aggregation}

\author{
Sungjae Lee$^{1,\dagger}$ \quad
Soheil Abbasloo$^{2,*}$ \quad
Jabin Koo$^{1}$ \quad
Jungseul Ok$^{1}$\\
$^{1}$POSTECH \quad $^{2}$Microsoft Research
}

\iclrfinalcopy
\begin{document}

\maketitle

\begingroup
\renewcommand{\thefootnote}{\fnsymbol{footnote}}
\footnotetext[2]{Work done during an internship at Microsoft Research Asia.}
\footnotetext[1]{Correspondence to: Soheil Abbasloo \texttt{\textless{}soheil.abbasloo@microsoft.com\textgreater{}}}
\endgroup

\begin{abstract}

Multi-agent reasoning with large language models (LLMs) raises a central question for collective decision-making: when agents disagree, whose decision should be trusted? Resolving such disagreements is difficult because agents' decisions are susceptible to non-semantic biases such as majority pressure and response-style preferences, which can obscure the quality of the underlying reasoning. Our analysis shows how bias can be both mitigated and exploited: \emph{bias control} improves initial answer selection, while \emph{bias injection} elicits decision instability that serves as a reliability signal. Building on these findings, we introduce a test-time aggregation framework that requires neither validation data nor prior knowledge of relative model strength. The framework implements bias control by presenting one style-normalized reasoning trace per distinct answer, and bias injection by placing each aggregating agent’s initially selected answer in the minority. Overall, our framework achieves higher average accuracy and better token efficiency than competing non-oracle aggregation methods across three mathematical reasoning benchmarks and four heterogeneous LLM combinations. It also improves final-answer accuracy when applied to multi-agent debate and Mixture-of-Agents, without modifying their interaction or synthesis processes.

\end{abstract}

\section{Introduction}
\label{sec:intro}

Multi-agent reasoning benefits from the diverse and complementary solutions produced by multiple agents, yet those agents often reach conflicting conclusions. Whether candidate solutions are generated independently, revised through multi-agent debate, or synthesized across multiple stages, a correct answer among these conflicting solutions may still be missed during final aggregation~\citep{ai2025beyond,du2023improving,wang2025mixture}. The effectiveness of multi-agent reasoning therefore depends not only on generating useful solutions, but also on reliably resolving these disagreements~\citep{wang2024rethinking,smit2023should,choi2026debate,wang2026slm}.

\begin{wrapfigure}[16]{r}{0.34\textwidth}
\vspace{-0.6\baselineskip}
\centering
\setlength{\abovecaptionskip}{3pt}
\setlength{\belowcaptionskip}{0pt}
\includegraphics[
    width=\linewidth,
    trim={0 2mm 0 2mm},
    clip
]{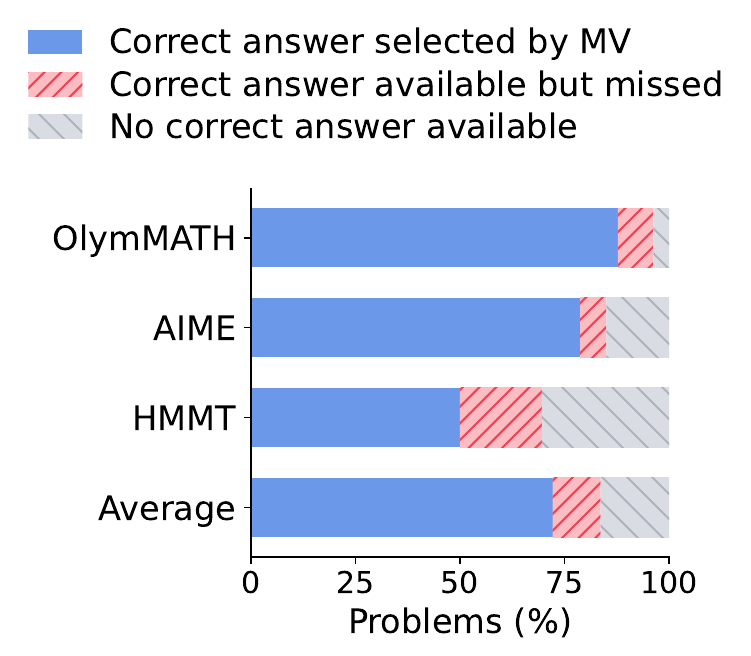}
\caption{\textbf{Correct-answer availability and majority-vote accuracy.}
Majority voting is effective overall but still overlooks correct answers with limited support.}
\label{fig:availability-gap}
\vspace{-0.5\baselineskip}
\end{wrapfigure}

To reach a collective decision, frequency- and consistency-based aggregation methods rely on agreement among responses as evidence of correctness~\citep{wang2022self,ai2025beyond,wang2026slm}. Agreement is often informative, but can favor a shared error over a correct minority answer. For example, majority voting (MV) is widely used and often effective. In our analysis, the sampled response pool contains a correct answer in 83.6\% of cases on average, yet MV selects it in only 72.2\% (Figure~\ref{fig:availability-gap}; see Section~\ref{sec:exp_setup} for details). Alternatively, LLM-based aggregation can consider the reasoning behind each answer as additional evidence, allowing a correct minority answer to prevail. However, incorporating reasoning does not by itself ensure better selection: LLM decisions can be distorted by frequency, style, position, and other non-semantic biases~\citep{ma2025judging,zhu2025conformity,choi2025empirical,wang2024large,shi2025judging,feuer2025style,hwang2025can,choi2026identity,chen2024humans}. How these biases are handled therefore matters for turning this additional reasoning evidence into better aggregation decisions.

\begin{figure*}[t]
    \centering
    \includegraphics[width=\textwidth]{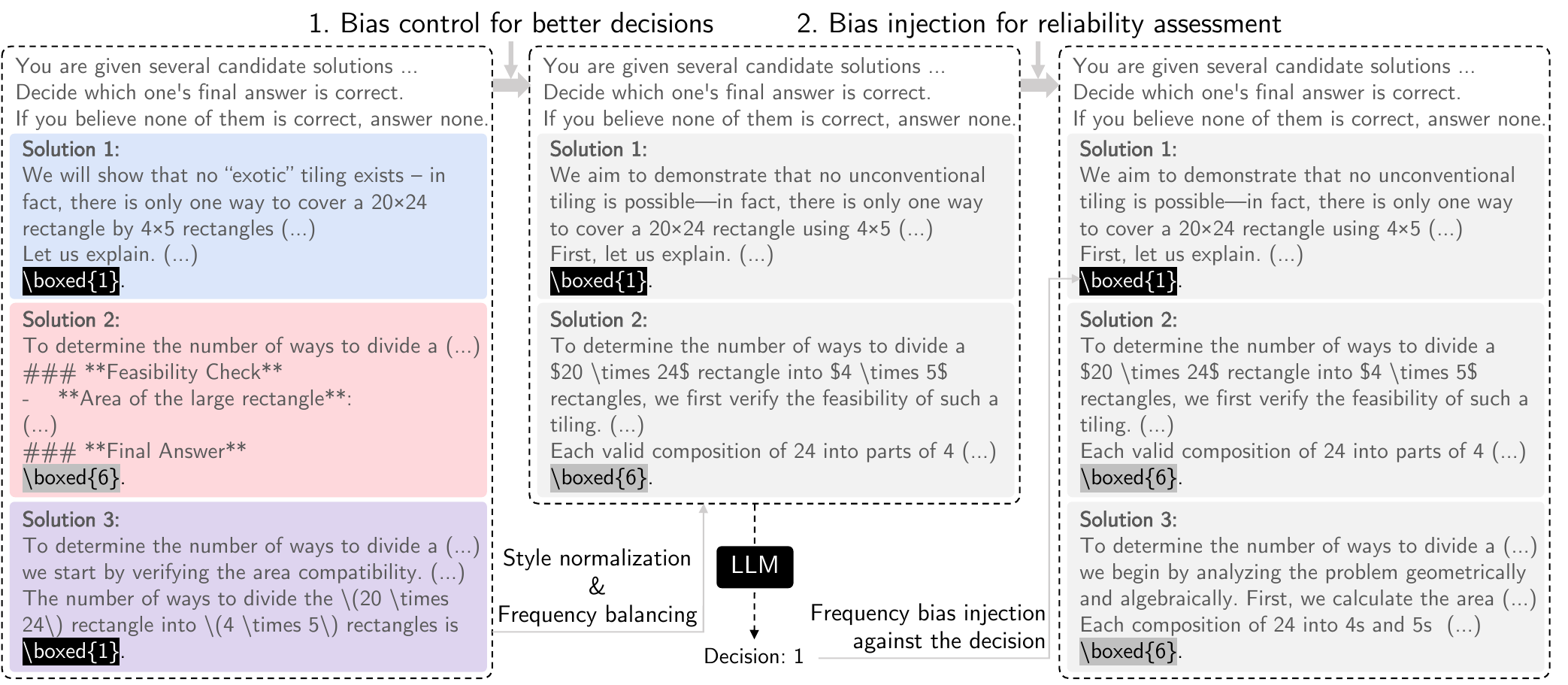}
    \vspace{-5mm}
    \caption{\textbf{Illustration of bias control and bias injection.}
    Bias control forms an initial decision by comparing one style-normalized trace per distinct answer under balanced frequency. Bias injection then places the selected answer in the relative minority to assess its reliability. Bias is mitigated through control during decision formation and then exploited through injection to assess reliability.}
    \label{fig:main-modules}
    \vspace{-5mm}
\end{figure*}

Most prior work seeks to mitigate biases in LLM decisions by reducing or controlling their influence~\citep{li2024split,dubois2024length,li2025calibraeval,yang2026any}. In contrast, our key idea is to both mitigate and exploit bias: we first control presentation bias to improve an initial decision, then deliberately inject bias against the selected answer to assess that decision's reliability. For bias control, we find that style normalization improves answer-selection accuracy beyond frequency balancing alone. For bias injection, we find that injecting frequency bias against the selected answer changes incorrect decisions more often than correct ones. The decision instability provides a more informative reliability signal than that obtained by reversing candidate
order or repeating the same query. Based on these findings, we introduce a test-time framework for multi-agent aggregation with two complementary operations, \emph{bias control} and \emph{bias injection} (Figure~\ref{fig:main-modules}).

The framework requires neither validation data nor prior knowledge of relative model strength. It operates directly on reasoning history, the collection of reasoning traces and their extracted answers available before final aggregation. Bias control retains one reasoning trace per distinct answer and rewrites each into a shared plain-prose form, helping the participating agents assess the reasoning with less influence from frequency and style biases. When their initial selections disagree, bias injection places each agent's selected answer in the minority to assess the reliability of that decision on the current problem. The resulting problem-specific reliability signal guides final-answer selection.

We evaluate the framework on independently sampled responses from heterogeneous LLMs across three mathematical reasoning benchmarks and four model combinations. Our framework achieves higher average accuracy and better token efficiency than competing non-oracle aggregation methods. Applying the same aggregation procedure to multi-agent debate~\citep{du2023improving} and Mixture-of-Agents (MoA)~\citep{wang2025mixture} also improves final-answer accuracy without modifying their interaction or synthesis processes.

\section{Related Work}

\paragraph{Multi-Agent Reasoning.}
Multi-agent reasoning seeks to improve LLM problem solving by combining the outputs of multiple agents~\citep{wang2026slm,tian2026multi,cui2026free}. Agents may generate solutions independently, exchange intermediate reasoning, critique one another, synthesize responses, or revise their answers over multiple rounds~\citep{du2023improving,wang2025mixture,choi2026debate,tian2026multi,cui2026free,zhu2026demystifying}. These approaches aim to increase reasoning diversity and combine complementary strengths across agents~\citep{wang2026slm,zhu2026demystifying}. In practice, however, interaction does not consistently provide such benefits~\citep{wang2024rethinking,smit2023should,choi2026debate,zhu2026demystifying}. Multi-agent interactions may converge to an incorrect consensus, amplify early errors, or encourage agents to follow a dominant opinion rather than assess the underlying reasoning~\citep{zhu2025conformity,choi2025empirical,okawaemergence,cui2026free}. As a result, a correct answer may appear among generated or intermediate responses without being selected as the final answer~\citep{cui2026free,wang2026slm}.

\paragraph{Final-Answer Aggregation.}
Multi-agent frameworks use different rules for final-answer aggregation. Multi-agent debate commonly relies on terminal-round responses or consensus~\citep{du2023improving,choi2026debate}; SLM-MUX selects among model-wise majority answers using within-model consistency, with ties broken by each model's historical accuracy measured on a separate validation set~\citep{wang2026slm}; and Mixture-of-Agents synthesizes responses across successive layers through a designated final aggregator~\citep{wang2025mixture}. 
However, majority voting and consistency-based aggregation can fail when model errors are correlated or a correct answer appears with low frequency~\citep{choi2026debate,ai2025beyond,cui2026free,wang2026can}. Recent work moves beyond raw answer counts by leveraging model accuracies and cross-model answer correlations~\citep{ai2025beyond}, scoring answers based on changes across debate rounds~\citep{cui2026free}, or retaining the best-supported rationale for each distinct answer~\citep{wang2026can}. Verification-based approaches similarly select among candidate answers using judgments from multiple verifiers~\citep{lifshitz2025multi}. Our framework draws candidate answers from the available reasoning history, which can include non-majority and intermediate answers. It assesses decision reliability independently for each problem without training, validation data, or prior knowledge of relative model strength.

\paragraph{Biases in Multi-Agent Aggregation.}
While reasoning traces provide additional evidence beyond final answers alone~\citep{abbasloo2025siev}, LLM interactions and judgments can be influenced by non-semantic cues such as answer frequency, prompt position, response style, and model identity~\citep{ma2025judging,zhu2025conformity,choi2025empirical,wang2024large,shi2025judging,feuer2025style,choi2026identity}. These effects also matter in heterogeneous systems, where different LLMs produce reasoning traces with different response styles~\citep{chen2025beyond,feuer2025style}. Consequently, a conclusion may be favored because it appears more frequently or is expressed in a preferred style rather than because its reasoning provides stronger evidence of correctness~\citep{ma2025judging,zhu2025conformity,choi2025empirical,hwang2025can,feuer2025style}. 
Most prior work focuses on analyzing biases in LLM decisions and mitigating their influence~\citep{ma2025judging,shi2025judging,li2024split,li2025calibraeval,ye2025justice,yang2026any,yang2026fairjudge,koo2026auditing}. Recent work also exploits biases to examine judge robustness~\citep{yang2026turning} and uses presentation perturbations to derive uncertainty signals~\citep{badshah2026scope}. We study the complementary use of bias mitigation through control and bias exploitation through injection for multi-agent aggregation.

\section{Motivation: Intervening on biases}
\label{sec:analysis}

We examine how to intervene on presentation biases to both mitigate and exploit their effects in multi-agent aggregation. Specifically, we ask whether controlling presentation biases improves agents' initial answer selection and whether injecting adverse bias against their selected answers provides a useful signal of decision reliability.

\paragraph{Analysis setup.}
To isolate final-answer decision from upstream interaction dynamics, we analyze independently sampled response pools for 220 mathematical reasoning problems from OlymMATH, AIME, and HMMT. For each problem, we target three completed reasoning traces from each of four heterogeneous reasoning models: Qwen3-8B (Qwen), Phi-4-reasoning-plus (Phi), GLM-Z1-9B (GLM), and Nemotron-Nano-9B-v2 (Nemotron). We consider all four three-model combinations and focus on instances not resolved by unanimous model-level agreement. The participating models also serve as judges over the competing answers. See Section~\ref{sec:experiments} for experimental details.

\subsection{Controlling Biases in Candidate Presentation}
\label{sec:analysis-control}

Following the one-rationale-per-answer construction of \citet{wang2026can}, we group reasoning traces by their final answers, retain one trace for each distinct answer, and present every answer once. The agent then selects among the resulting candidates or \texttt{none} if none is correct. This removes the original majority--minority relation and yields a frequency-balanced comparison.

Frequency balancing controls count-based cues, but not the source-dependent presentation patterns naturally embedded in reasoning traces generated by different LLMs. To control this remaining variation rather than compare raw traces directly, we rewrite each retained trace with the same prompt into a common plain-prose form. The prompt standardizes surface wording and formatting while instructing the rewriter to preserve the original length, reasoning content, numerical values, equations, and final answer. The normalization prompt and examples of original and normalized traces are provided in Figure~\ref{fig:appendix-prompt-normalization} and Appendix~\ref{app:normalization-examples}, respectively.

\begin{wraptable}[15]{r}{0.62\textwidth}
\centering
\vspace{-0.7\baselineskip}
\footnotesize
\setlength{\tabcolsep}{2.5pt}
\renewcommand{\arraystretch}{1.0}

\caption{\textbf{Effect of style normalization on LLM decision accuracy.}
Rows identify the models forming the candidate pool, and columns identify the agent making the decision. Cells report Original $\to$ Normalized accuracy under frequency-balanced presentation on paired instances valid in both conditions; bold marks the higher value. Style normalization generally improves accuracy.}

\begin{adjustbox}{max width=\linewidth}
\begin{tabular}{@{}lcccc@{}}
\toprule
\multirow{2}{*}{Models}
& \multicolumn{4}{c}{Answer-selection accuracy by agent (\%)} \\
\cmidrule(lr){2-5}
& Qwen & Phi & GLM & Nemo \\
\midrule

Qwen--Phi--GLM
& 60.8$\to$60.8
& 65.0$\to$\textbf{70.7}
& 62.4$\to$\textbf{63.8}
& -- \\

Qwen--Phi--Nemotron
& 62.1$\to$\textbf{63.9}
& 68.2$\to$\textbf{70.6}
& --
& 57.8$\to$\textbf{58.1} \\

Qwen--GLM--Nemotron
& \textbf{55.7}$\to$54.8
& --
& 54.2$\to$\textbf{59.4}
& 50.4$\to$\textbf{51.9} \\

Phi--GLM--Nemotron
& --
& \textbf{65.2}$\to$64.2
& 65.1$\to$\textbf{67.9}
& \textbf{63.7}$\to$60.9 \\

\bottomrule
\end{tabular}
\end{adjustbox}

\label{tab:style-parseable}
\vspace{-0.5\baselineskip}
\end{wraptable}

We compare paired decisions over the same problems, candidate answers, and source traces, with the normalized condition presenting the rewritten traces. To distinguish answer selection from parsing failures, we include only decision pairs whose selected answers can be successfully parsed in both conditions. As shown in Table~\ref{tab:style-parseable}, style normalization improves pooled answer-selection accuracy and maintains or improves accuracy in 9 of the 12 agent--combination settings. These conditional results indicate that style normalization can improve decision accuracy beyond frequency balancing.

\subsection{Injecting Bias to Elicit Decision Instability}
\label{sec:analysis-probing}

Answer-frequency and conformity effects have been repeatedly observed in LLM decisions~\citep{ma2025judging,zhu2025conformity,choi2025empirical}, motivating their control during decision formation in Section~\ref{sec:analysis-control}. We ask whether this same susceptibility can instead be exploited after an initial controlled decision to assess its reliability through decision instability. We define decision instability as a change between the initial controlled decision and the second-pass decision. Placing the selected answer at a presentation disadvantage creates a stress test: a decision grounded in the underlying reasoning should be more likely to persist than a less reliable one. Here, the signal is whether the initial decision changes, not whether the perturbed query itself produces a better answer.

\begin{wrapfigure}{r}{0.42\textwidth}
    \vspace{-0.6\baselineskip}
    \centering
    \includegraphics[width=\linewidth]{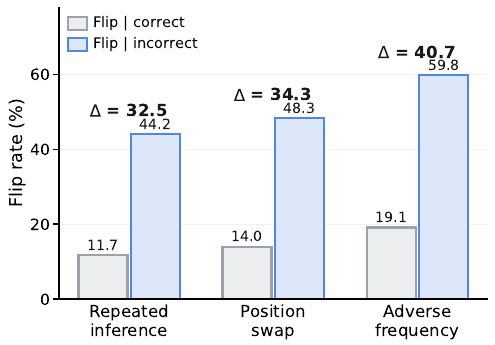}
    \vspace{-5mm}
    \caption{\textbf{Decision instability under bias injection.}
    Bars show flip rates conditioned on initial correctness. Adverse-frequency yields the largest gap, offering a more selective signal for reliability assessment.}
    \label{fig:bias-probes}
    \vspace{-5mm}
\end{wrapfigure}

Holding the set of distinct answer hypotheses fixed, we compare three second-pass conditions. Repeated inference queries the judge again with the same bias-controlled presentation. Position swap reverses the candidate order, following prior work that uses bidirectional response orderings to derive uncertainty signals for selective judging~\citep{badshah2026scope}. Adverse-frequency injection applies stronger pressure by presenting the agent's initially selected answer once and its strongest competing answer twice, placing the initial choice in the relative minority.

A useful second-pass intervention for final aggregation should change incorrect decisions more often than correct ones. As shown in Figure~\ref{fig:bias-probes}, adverse-frequency injection changes 59.8\% of incorrect decisions, compared with 19.1\% of correct decisions, yielding a 40.7-point gap. This exceeds the corresponding gaps of 32.5 points for repeated inference and 34.3 points for position swap, indicating that bias-induced decision instability provides a more selective reliability signal for aggregation. Section~\ref{sec:analysis_probe} shows that a substantial selectivity gap remains when the added trace is an exact copy. 

\begin{figure*}[t]
    \centering
    \includegraphics[width=\textwidth]
    {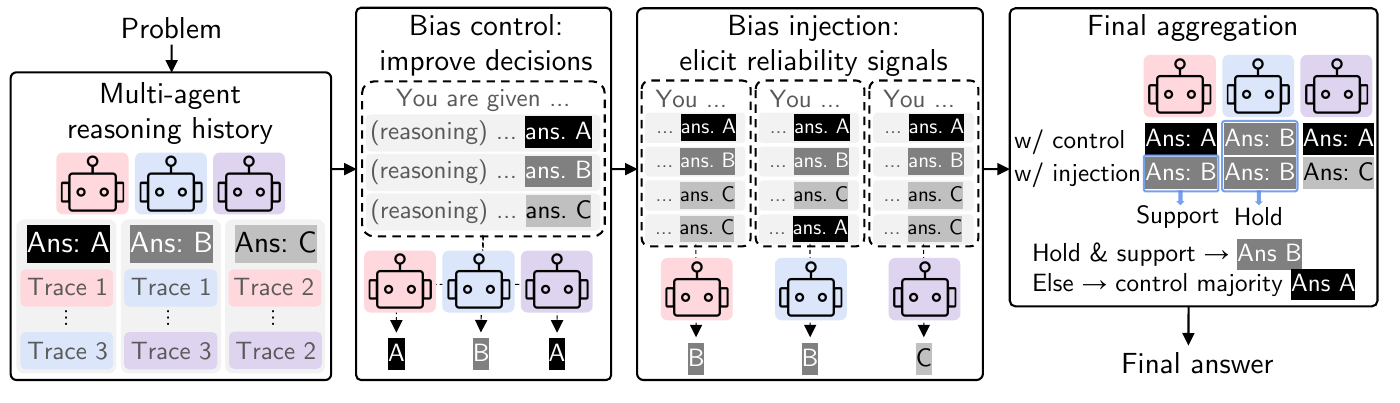}
    \vspace{-5mm}
    \caption{\textbf{Overview of the proposed aggregation framework.}
    Responses from the reasoning history are grouped by answer. When agents disagree, the framework uses bias control to form LLM decisions and injects adverse-frequency bias to test their reliability. Final aggregation combines decision holds observed under bias injection with cross-agent support, without modifying upstream reasoning.}
    \label{fig:main-framework}
    \vspace{-5mm}
\end{figure*}

\section{Proposed Method}
\label{sec:method}

Building on the findings in Section~\ref{sec:analysis}, we introduce a training- and validation-free aggregation framework (Figure~\ref{fig:main-framework}). Cross-model agreement and disagreement provide useful information about decision uncertainty~\citep{hamidieh2026complementing}. We therefore apply an upstream agreement gate, returning the agreed answer when it is satisfied and invoking additional aggregation otherwise. For independently sampled responses, the gate is satisfied when every participating model has the same unique modal answer among its completed responses; a within-model tie does not satisfy the gate. Bias control forms initial decisions; disagreement among these decisions triggers bias injection, which uses decision instability to assess their reliability on the current problem. The framework requires neither prior estimates of model strength nor a designated final aggregator and applies to independently sampled responses, multi-agent debate~\citep{du2023improving}, and Mixture-of-Agents (MoA)~\citep{wang2025mixture}, without modifying upstream reasoning.

\paragraph{Bias control for initial decision.}
To improve initial answer selection, we place distinct answer hypotheses under more comparable presentation conditions. Given a problem and its upstream reasoning history, we group reasoning traces by their extracted final answers and retain one trace for each distinct answer. We present each retained answer once to balance its presentation frequency and rewrite each retained trace into a common plain-prose form to reduce source-dependent stylistic differences. We submit the resulting candidate set to $J$ aggregating agents and let $c_j$ denote the answer selected by agent $j$. If the aggregating agents unanimously select the same answer after bias control, we return it; otherwise, we proceed to bias injection.

\paragraph{Bias injection for decision instability assessment.}
Resolving disagreement requires problem-specific evidence about which decisions remain reliable. Rather than assigning fixed reliability to agents, we stress-test each controlled decision on the current problem and observe whether it changes under an adverse presentation. For each agent, we construct a second-pass input based on its controlled decision. If the reasoning history contains an additional trace for the competing answer with the greatest original support, we apply adverse-frequency injection: the initially selected answer is presented once, whereas the competing answer is presented twice using distinct normalized traces. This places the initial choice in the relative minority while leaving the set of answer hypotheses unchanged. If no such additional trace is available, we instead reverse the candidate order. Let $p_j$ denote agent $j$'s second-pass decision. The complete decision prompts are provided in Figure~\ref{fig:appendix-prompt-judgment-frequency}.

\paragraph{Final aggregation.}
To combine decisions from bias control and bias injection, we define the hold count $H(a)=|\{j:c_j=p_j=a\}|$ and second-pass support count $S(a)=|\{j:p_j=a\}|$ for each answer $a$. We select $a$ if at least one agent retains it and a majority supports it on the second pass, i.e., $H(a)>0$ and $S(a)>J/2$. If no answer satisfies this condition, we return the strict-majority answer among the controlled decisions when one exists; otherwise, we select the answer with the most votes across both stages. Original response frequencies inform candidate construction but are not counted as votes in the final decision. Algorithm~\ref{alg:ours} provides the complete aggregation procedure.

\section{Experiments}
\label{sec:experiments}

\subsection{Experimental Setup}
\label{sec:exp_setup}

\paragraph{Models.}
We evaluate four open-weight reasoning models: Qwen3-8B~\citep{yang2025qwen3}, Phi-4-reasoning-plus (14B)~\citep{abdin2025phi}, GLM-Z1-9B-0414~\citep{glm2024chatglm}, and NVIDIA-Nemotron-Nano-9B-v2~\citep{basant2025nvidia}, referred to as Qwen, Phi, GLM, and Nemotron, respectively. We consider all four three-model combinations: Qwen--Phi--GLM, Qwen--Phi--Nemotron, Qwen--GLM--Nemotron, and Phi--GLM--Nemotron. We use Qwen as the default style normalizer; Appendix~\ref{app:normalizer-robustness} shows comparable final accuracy with different normalizers. We use the official model checkpoints and their default reasoning modes; full inference details are provided in Appendix~\ref{app:implementation}.

\paragraph{Benchmarks.}
We evaluate mathematical reasoning on 220 problems from three benchmarks: 100 problems from the English easy split of OlymMATH~\citep{sun2026challenging}, 60 from AIME 2024--2025~\citep{aime}, and 60 from the February HMMT 2024--2025 competitions~\citep{hmmt}. We report final-answer accuracy using the same answer normalization and equivalence checks across all benchmarks and aggregation methods.

\paragraph{Baselines.}
\emph{Combined MV} selects the most frequent answer across all sampled responses. \emph{SLM-MUX}~\citep{wang2026slm} selects among model-wise modal answers using within-model consistency. Because our setting fixes model combinations and assumes no validation data, we omit its validation-based subset search and replace its validation-accuracy tie-breaker with plurality voting over the tied models' completed samples, breaking residual ties by first occurrence. Appendix~\ref{app:slm-mux-tiebreaking} evaluates a test-set-informed oracle tie-breaker; Ours achieves higher average accuracy under either tie-breaking rule. \emph{Multi-Agent Verification} (MAV)~\citep{lifshitz2025multi} applies aspect-specific binary verification to the same candidate answers as our method, using one representative trace per answer. For each selected aspect, all three participating models evaluate every candidate, and we select the candidate with the most approvals. We also report \emph{Single-Best-SC}~\citep{wang2022self}, which retrospectively selects the strongest individual model under self-consistency and serves as an oracle model-selection reference rather than a deployable baseline.

\paragraph{Evaluation protocol.}
Our primary metric is final-answer accuracy, and we use mean total token usage, including input and output tokens, per problem as a proxy for inference cost. Methods with LLM-based aggregation are run three times over the fixed response pools, and their accuracies are averaged; rule-based methods are evaluated once. For MAV, we fix $k=3$ and cumulatively include answer correctness, reasoning validity, and solution completeness to obtain three accuracy--cost points. For MoA, we evaluate each participating model as the final aggregator and retrospectively report the best-performing choice as an oracle reference. We target $k$ completed responses per model, allowing at most $2k$ generation attempts and including tokens from all attempts in the reported cost. Our primary setting uses $k=3$, with additional response budgets used to characterize the accuracy--cost trade-off. In this setting, the agreement-routing rule in Section~\ref{sec:method} operates on the model-wise majority answers over the completed responses. In our framework, we retain up to four distinct candidate answers with the highest frequencies in the reasoning history and present them in descending frequency order.

\subsection{Experimental Results}
\label{sec:results}

\begin{figure*}[t]
\centering
\includegraphics[width=\textwidth]{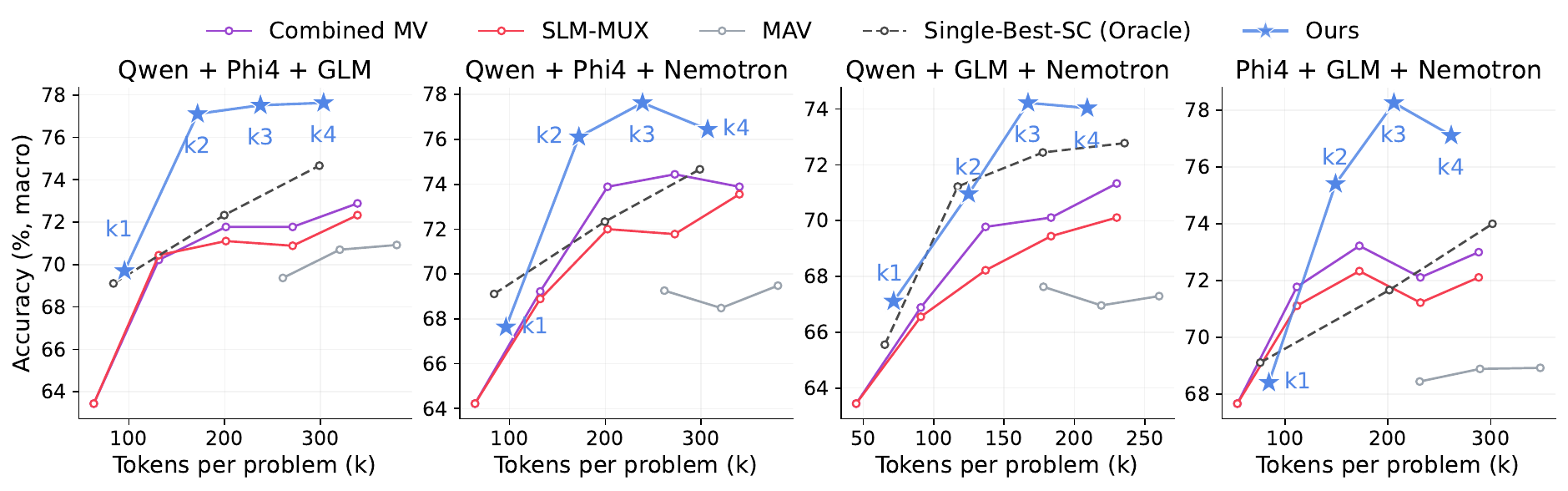}
\vspace{-6mm}
\caption{\textbf{Accuracy--cost trade-offs under independent sampling.}
Each panel shows one three-model combination across response budgets. Accuracy is the macro-average across OlymMATH, AIME, and HMMT, and inference cost is measured by total input-and-output tokens per problem. Stars mark Ours. Most Ours configurations with $k\in\{2,3,4\}$ lie on or near the observed Pareto frontier.}
\label{fig:cost-accuracy}
\end{figure*}

\begin{table*}[t!]
\centering
\footnotesize
\setlength{\tabcolsep}{3.0pt}
\renewcommand{\arraystretch}{0.96}

\caption{\textbf{Final-answer accuracy and inference efficiency across heterogeneous LLM combinations.}
Each baseline uses the nearest evaluated token budget at or above Ours. Average is the unweighted benchmark mean, and Tokens denotes input-and-output usage per problem. For LLM-based aggregation methods, values are mean $\pm$ standard deviation over three runs. Single-Best-SC is an oracle model-selection reference. Bold and underline mark the first- and second-ranked results, respectively. Ours achieves the highest average accuracy with fewer tokens in every combination.}
\label{tab:main-results}

\begin{adjustbox}{width=\textwidth}
\begin{tabular}{clccccc}
\toprule
Model combination & Method & OlymMATH $\uparrow$ & AIME $\uparrow$ & HMMT $\uparrow$ & Average $\uparrow$ & Tokens ($10^3$) $\downarrow$ \\
\midrule

\multirow{5}{*}{Qwen--Phi--GLM}
& Single-Best-SC~\citep{wang2022self} & \underline{89.0} & \textbf{83.3} & 51.7 & \underline{74.7} & 298.7 \\
\cmidrule(lr){2-7}
& Combined MV & 87.0 & \underline{80.0} & 48.3 & 71.8 & 270.8 \\
& MAV~\citep{lifshitz2025multi} & 82.0{\scriptsize$\pm$2.0} & 71.1{\scriptsize$\pm$2.5} & \underline{55.0}{\scriptsize$\pm$2.9} & 69.4{\scriptsize$\pm$0.9} & \underline{260.6}{\scriptsize$\pm$0.9} \\
& SLM-MUX~\citep{wang2026slm} & 86.0 & 78.3 & 48.3 & 70.9 & 270.8 \\
& \ourscell{Ours} & \ourscell{\textbf{90.3}{\scriptsize$\pm$1.2}} & \ourscell{\textbf{83.3}{\scriptsize$\pm$1.7}} & \ourscell{\textbf{58.9}{\scriptsize$\pm$1.9}} & \ourscell{\textbf{77.5}{\scriptsize$\pm$1.5}} & \ourscell{\textbf{237.3}{\scriptsize$\pm$1.2}} \\
\midrule

\multirow{5}{*}{Qwen--Phi--Nemotron}
& Single-Best-SC & 89.0 & \textbf{83.3} & 51.7 & \underline{74.7} & 298.7 \\
\cmidrule(lr){2-7}
& Combined MV & \underline{90.0} & 78.3 & 55.0 & 74.4 & 272.5 \\
& MAV & 81.7{\scriptsize$\pm$0.6} & 70.6{\scriptsize$\pm$1.0} & \underline{55.6}{\scriptsize$\pm$4.2} & 69.3{\scriptsize$\pm$1.2} & \underline{261.6}{\scriptsize$\pm$0.9} \\
& SLM-MUX & 87.0 & 76.7 & 51.7 & 71.8 & 272.5 \\
& \ourscell{Ours} & \ourscell{\textbf{90.7}{\scriptsize$\pm$0.6}} & \ourscell{\underline{82.8}{\scriptsize$\pm$1.0}} & \ourscell{\textbf{59.4}{\scriptsize$\pm$1.9}} & \ourscell{\textbf{77.6}{\scriptsize$\pm$0.1}} & \ourscell{\textbf{238.9}{\scriptsize$\pm$1.0}} \\
\midrule

\multirow{5}{*}{Qwen--GLM--Nemotron}
& Single-Best-SC & \underline{89.0} & \textbf{80.0} & 48.3 & \underline{72.4} & \underline{177.6} \\
\cmidrule(lr){2-7}
& Combined MV & 87.0 & 76.7 & 46.7 & 70.1 & 183.4 \\
& MAV & 77.3{\scriptsize$\pm$3.2} & 75.6{\scriptsize$\pm$2.5} & \underline{50.0}{\scriptsize$\pm$2.9} & 67.6{\scriptsize$\pm$2.0} & 178.0{\scriptsize$\pm$0.6} \\
& SLM-MUX & 85.0 & 76.7 & 46.7 & 69.4 & 183.4 \\
& \ourscell{Ours} & \ourscell{\textbf{89.3}{\scriptsize$\pm$1.5}} & \ourscell{\underline{78.9}{\scriptsize$\pm$1.9}} & \ourscell{\textbf{54.4}{\scriptsize$\pm$1.0}} & \ourscell{\textbf{74.2}{\scriptsize$\pm$0.6}} & \ourscell{\textbf{167.1}{\scriptsize$\pm$1.0}} \\
\midrule

\multirow{5}{*}{Phi--GLM--Nemotron}
& Single-Best-SC & 87.0 & \underline{83.3} & \underline{51.7} & \underline{74.0} & 301.7 \\
\cmidrule(lr){2-7}
& Combined MV & \underline{88.0} & 80.0 & 48.3 & 72.1 & 231.7 \\
& MAV & 82.0{\scriptsize$\pm$1.0} & 71.7{\scriptsize$\pm$2.9} & \underline{51.7}{\scriptsize$\pm$2.9} & 68.4{\scriptsize$\pm$2.0} & \underline{231.0}{\scriptsize$\pm$0.4} \\
& SLM-MUX & 87.0 & 78.3 & 48.3 & 71.2 & 231.7 \\
& \ourscell{Ours} & \ourscell{\textbf{88.7}{\scriptsize$\pm$0.6}} & \ourscell{\textbf{85.0}{\scriptsize$\pm$0.0}} & \ourscell{\textbf{61.1}{\scriptsize$\pm$1.0}} & \ourscell{\textbf{78.3}{\scriptsize$\pm$0.5}} & \ourscell{\textbf{206.0}{\scriptsize$\pm$2.0}} \\
\bottomrule

\end{tabular}
\end{adjustbox}
\end{table*}

\paragraph{Main results.}

For independently generated multi-LLM reasoning, our primary comparison examines the accuracy--cost trade-off across response budgets (Figure~\ref{fig:cost-accuracy}). Across all four model combinations, most Ours configurations with $k\in\{2,3,4\}$ lie on or near the observed Pareto frontier. Table~\ref{tab:main-results} complements this comparison under our budget-matching protocol, reporting mean $\pm$ standard deviation over three runs for LLM-based aggregation methods. Here, Ours achieves the highest average accuracy in all four model combinations, outperforming the strongest non-oracle baseline by 3.2--6.2 percentage points and the retrospective Single-Best-SC oracle reference by 1.8--4.3 points. It also achieves the best or tied-best non-oracle accuracy in 12 of the 12 benchmark--combination pairs. We additionally evaluate a common subset of instances not resolved by the upstream agreement gate, determined from the $k=3$ response pools before aggregation. Applying the same budget-matching rule within this subset, Ours outperforms the strongest non-oracle baseline by 5.8--9.5 percentage points across model combinations (Appendix~\ref{app:upstream-disagreement}).

\paragraph{Recovery and retention relative to majority voting.}

\begin{wrapfigure}[11]{r}{0.60\textwidth}
\centering
\vspace{-0.5\baselineskip}
\setlength{\abovecaptionskip}{3pt}
\setlength{\belowcaptionskip}{0pt}
\includegraphics[width=\linewidth]{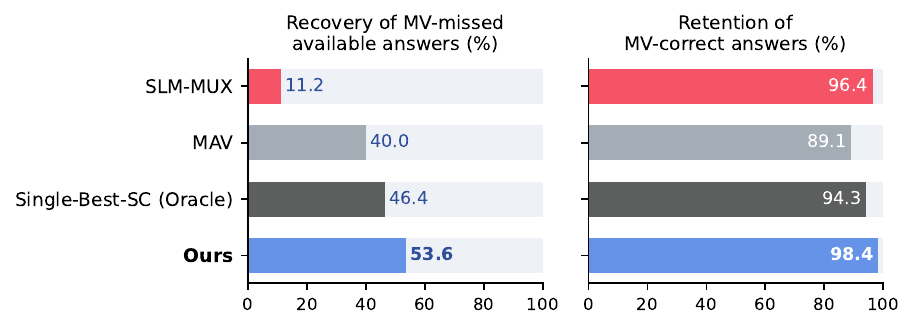}
\captionsetup{font=footnotesize}
\caption{\textbf{Recovery and retention relative to majority voting.} Rates are macro-averaged across benchmarks. Ours achieves the highest recovery while retaining most MV-correct answers.}
\label{fig:recovery-retention}
\vspace{-0.5\baselineskip}
\end{wrapfigure}

We examine whether aggregation recovers overlooked correct answers while preserving correct selections. Using the same $k=3$ reference response pools as Figure~\ref{fig:availability-gap}, we define recovery as accuracy when a correct answer is available but Combined MV is incorrect, and retention as accuracy when Combined MV is correct. As shown in Figure~\ref{fig:recovery-retention}, Ours achieves both the highest recovery rate, at 53.6\%, and the highest retention rate, at 98.4\%. MAV recovers 40.0\% of missed answers with 89.1\% retention, whereas SLM-MUX retains 96.4\% but recovers only 11.2\%. Single-Best-SC, included as an oracle reference with a target of nine completed responses from the retrospectively strongest model, achieves 46.4\% recovery and 94.3\% retention. These results show that Ours recovers more available correct answers while preserving nearly all MV-correct selections.

\paragraph{Application to multi-agent reasoning.}

\begin{wrapfigure}[16]{r}{0.40\textwidth}
\vspace{-0.6\baselineskip}
\centering
\setlength{\abovecaptionskip}{3pt}
\setlength{\belowcaptionskip}{0pt}
\includegraphics[width=\linewidth]
{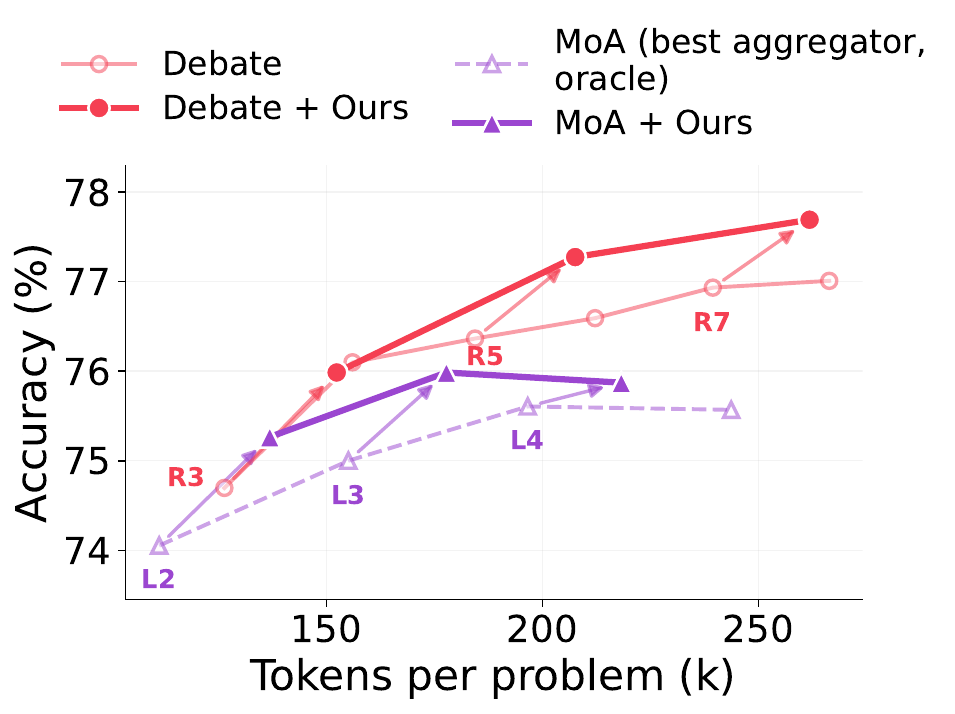}
\caption{\textbf{Integration with multi-agent reasoning.}
Arrows show pooled-accuracy gains from adding Ours.
R/L denote Debate rounds/MoA layers.}
\label{fig:integration}
\end{wrapfigure}

We next apply the same decision procedure to reasoning histories produced by
different multi-agent interaction frameworks such as \emph{multi-agent debate} (Debate)~\citep{du2023improving} and \emph{Mixture-of-Agents} (MoA)~\citep{wang2025mixture}. For each configuration, we keep the upstream reasoning history fixed and compare the native final decision with the corresponding \textsc{+Ours} variant. Since interaction may lead to biased consensus, we apply \textsc{+Ours} only when final answers disagree, their shared answer is supported by fewer than two initial, pre-interaction responses, or, for Debate, any agent changes its answer between the last two rounds. For MoA, agreement is checked among all final-layer proposers and the selected aggregator. Figure~\ref{fig:integration} shows that our method improves final-answer accuracy for both frameworks at a moderate additional inference cost.

\section{Analysis}
\label{sec:further-analysis}

\begin{wrapfigure}{r}{0.30\textwidth}
\centering
\vspace{-0.5\baselineskip}
\includegraphics[width=0.88\linewidth]
{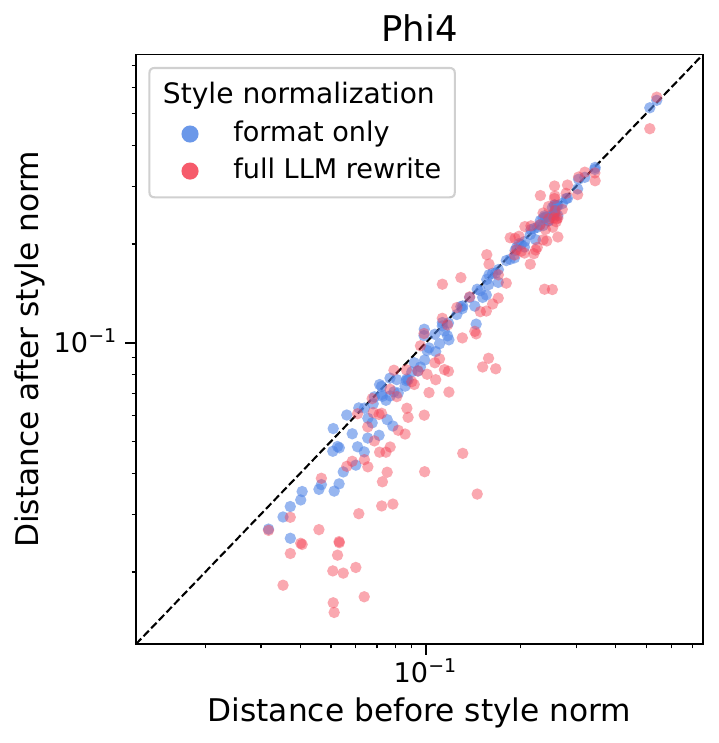}
\captionsetup{font=footnotesize,skip=3pt}
\caption{\textbf{Cross-generator convergence in Phi.} Cosine distances between matched traces.}
\label{fig:style_convergence}
\end{wrapfigure}

\subsection{Understanding Style Normalization}
\label{sec:style_analysis}

\paragraph{More than formatting removal.}
We examine whether the cross-generator convergence induced by style normalization can be explained by formatting removal alone. We construct matched pairs of reasoning traces produced by different generators for the same problem and answer, and compare deterministic formatting removal with full normalization. Figure~\ref{fig:style_convergence} shows the comparison in the Phi representation space, which is not used for rewriting. Formatting removal reduces mean representation distance by only 1.7\%, whereas full normalization reduces it by 8.2\%. Results in the representation spaces of three LLMs are provided in Appendix~\ref{app:style_convergence}.
Additional source-attribution analyses assess how accurately a trace's generating model can be identified from presentation features. Normalization reduces identification accuracy, particularly for formatting features, suggesting weaker model-specific presentation cues (Appendix~\ref{app:source-attribution}).

\subsection{Understanding Bias Injection}
\label{sec:analysis_probe}

\begin{wraptable}{r}{0.45\textwidth}
\centering
\vspace{-0.5\baselineskip}
\fontsize{8}{9}\selectfont
\setlength{\tabcolsep}{3.5pt}
\renewcommand{\arraystretch}{1.05}

\captionsetup{font=footnotesize}
\caption{\textbf{Reliability of retained decisions.}
Hold rate is the fraction unchanged on the second pass;
held accuracy is their conditional accuracy.
$\Delta$ is the gain over the 30.3\% controlled baseline.}
\label{tab:probe_filter}
\vspace{-2mm}

\begin{tabular}{@{}lccc@{}}
\toprule
Condition
& \shortstack{Hold rate\\(\%)}
& \shortstack{Held acc.\\(\%)}
& \shortstack{$\Delta$\\(pp)} \\
\midrule
Repeated inference & 53.7 & 40.6 & $+10.3$ \\
Position swap      & 52.4 & 42.4 & $+12.1$ \\
Adverse frequency  & 37.5 & 46.9 & $+16.6$ \\
\bottomrule
\end{tabular}

\vspace{-0.5\baselineskip}
\end{wraptable}

\paragraph{Does bias injection yield a more reliable retained set?}
Section~\ref{sec:analysis-probing} showed that adverse-frequency bias injection dislodges incorrect decisions more selectively than repeated inference or position swap. We analyze instances where the controlled decisions are not unanimous and a second pass is issued, comparing the reliability of decisions that remain unchanged under each second-pass condition. 
As shown in Table~\ref{tab:probe_filter}, adverse-frequency injection retains 37.5\% of controlled decisions, whose conditional accuracy is 46.9\%, compared with 42.4\% under position swap and 40.6\% under repeated inference. Relative to the common controlled-decision accuracy of 30.3\%, adverse-frequency injection yields the largest gain in retained-set accuracy, at 16.6 percentage points. It thus retains fewer decisions but produces a more accurate retained set, supporting decision persistence as a reliability signal. Additional analysis of final aggregation is provided in Appendix~\ref{app:final_aggregation}.

\begin{wraptable}[9]{r}{0.55\textwidth}
\centering
\vspace{-0.5\baselineskip}
\fontsize{8}{9}\selectfont

\setlength{\tabcolsep}{3.5pt}
\renewcommand{\arraystretch}{1.05}

\captionsetup{font=footnotesize}
\caption{\textbf{Content control for bias injection.}
Flip rates (\%) are conditioned on initial correctness.
Gap is the incorrect--correct difference in percentage points.}
\label{tab:frequency-copy-control}
\vspace{-2mm}

\begin{tabular}{@{}lccc@{}}
\toprule
Condition & Flip $\mid$ incorrect & Flip $\mid$ correct & Gap \\
\midrule
Repeated inference & 44.2 & 11.7 & 32.5 \\
Position swap      & 48.3 & 14.0 & 34.3 \\
Adverse freq. (exact copy) & 56.2 & 18.9 & 37.3 \\
Adverse freq. (distinct)   & 59.8 & 19.1 & 40.7 \\
\bottomrule
\end{tabular}

\vspace{-0.5\baselineskip}
\end{wraptable}

\paragraph{Does selectivity persist without new reasoning content?}
Our default injection adds a distinct trace for a competing answer, changing both presentation frequency and reasoning content. To examine the role of additional reasoning content, we compare second-pass conditions on instances with upstream disagreement, using the same initial controlled decisions across conditions. The exact-copy condition duplicates the competing trace already present, holding the competing answer, presentation order, and number of input blocks fixed. As shown in Table~\ref{tab:frequency-copy-control}, exact-copy injection yields an incorrect--correct flip gap of 37.3 percentage points, compared with 32.5 for repeated inference and 34.3 for position swap. Using a distinct trace yields the largest observed gap, at 40.7 points. These results suggest a role for frequency bias in selectivity and support using distinct reasoning traces when available.

\subsection{Ablation}
\label{sec:ablation}

\begin{wrapfigure}[15]{r}{0.43\textwidth}
\centering
\vspace{-0.6\baselineskip}
\includegraphics[width=\linewidth]{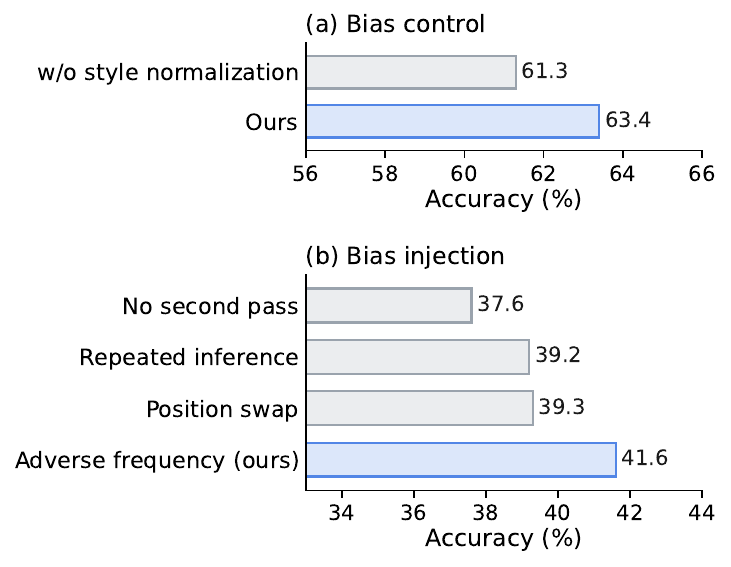}
\captionsetup{font=footnotesize,skip=3pt}
\caption{\textbf{Component ablations.}
Answer accuracy for (a) bias control and (b) bias injection.}
\label{fig:ablation}
\vspace{-0.5\baselineskip}
\end{wrapfigure}

Figure~\ref{fig:ablation} compares interventions on common evaluation sets comprising all instances reaching each stage. Accuracy is pooled across four model combinations and three runs. For bias control, we evaluate instances unresolved by the initial agreement gate, since normalization can affect subsequent controlled decisions and their agreement. Removing normalization reduces accuracy from 63.4\% to 61.3\%. For bias injection, we fix controlled decisions and compare second-pass strategies on instances where they disagree. The second-pass variants use the same final aggregation rule, while the no-second-pass baseline uses controlled decisions alone. Adverse-frequency injection achieves 41.6\% accuracy, compared with 37.6\% without a second pass, 39.2\% with repeated inference, and 39.3\% with position swap. Among strategies using the same number of second-pass decisions, adverse-frequency injection yields the highest accuracy under the shared aggregation rule.

\section{Conclusions}
\label{sec:conclusions}

We introduced a training- and validation-free framework that both mitigates and exploits biases in multi-agent aggregation. Bias control reduces the influence of frequency and style to improve initial decisions, while bias injection elicits decision instability as a problem-specific reliability signal. Across three mathematical reasoning benchmarks and four heterogeneous model combinations, the framework improves accuracy and token efficiency. It also improves final-answer aggregation in debate and MoA without modifying their upstream processes.

\section{Limitations}
\label{sec:limitations}

First, our evaluation focuses on mathematical reasoning with four open-weight models in the 8B--14B range. Broader evaluation would help assess how the benefits of bias control and the reliability signal from bias injection extend to other domains and larger models. Second, we study bias interventions at final aggregation, including for Debate and MoA, while keeping upstream reasoning histories fixed. Extending these interventions to interaction or synthesis is a direction for future work, with potential effects on both solution diversity and how agents reach agreement.

\section*{Reproducibility Statement}
The aggregation algorithm, experimental settings, and prompts are provided in Appendices~\ref{app:general-algorithm}, \ref{app:implementation}, and \ref{sec:app_prompts}, respectively. We will also release the source code and scripts needed to reproduce the reported experiments.

\section*{AI Use Statement}
Generative AI assistants were used for manuscript editing and formatting, and for assistance with experimental and analysis code. All AI-assisted outputs were reviewed and verified by the authors, who take full responsibility for the final content of this work.

\bibliography{iclr2027_conference}
\bibliographystyle{iclr2027_conference}

\clearpage

\appendix

\section{Aggregation Algorithm}
\label{app:general-algorithm}

Algorithm~\ref{alg:ours} summarizes the complete aggregation procedure.

\begin{algorithm}[H]
\caption{Bias Control and Bias Injection for Multi-Agent Aggregation}
\label{alg:ours}
\begin{algorithmic}[1]
    \STATE \textbf{Input:} Problem $q$, reasoning history
    $\mathcal{H}=\{(r_i,a_i)\}_{i=1}^{N}$, upstream conclusions
    $\mathcal{U}$, judges $\{\mathcal{J}_j\}_{j=1}^{J}$, normalizer
    $G_{\mathrm{norm}}$
    \STATE \textbf{Output:} Final answer $\widehat{a}$

    \hfill \textbf{Step 1: Agreement Routing}

    \IF{the upstream agreement gate is satisfied}
        \STATE \textbf{return} the agreed answer
    \ENDIF

    \hfill \textbf{Step 2: Bias Control for Decision Formation}

    \STATE Group $\mathcal{H}$ by answer into
    $\{\mathcal{R}_a\}_{a\in\mathcal{A}}$ and set
    $n(a)\leftarrow|\mathcal{R}_a|$
    \STATE Select one $r_a\in\mathcal{R}_a$ and set
    $\widetilde r_a\leftarrow G_{\mathrm{norm}}(r_a)$
    for each $a\in\mathcal{A}$
    \STATE $\mathcal{S}^{\mathrm{ctrl}}
    \leftarrow[(\widetilde r_a,a)]_{a\in\mathcal{A}}$,
    ordered by decreasing $n(a)$
    \STATE $c_j\leftarrow
    \mathcal{J}_j(q,\mathcal{S}^{\mathrm{ctrl}})$
    for $j=1,\ldots,J$
    \IF{$c_1=\cdots=c_J$}
        \STATE \textbf{return} $c_1$
    \ENDIF

    \hfill \textbf{Step 3: Bias Injection for Reliability Assessment}

    \FOR{$j=1,\ldots,J$}
        \STATE $b_j\leftarrow
        \operatorname*{arg\,max}_{a\in\mathcal{A}\setminus\{c_j\}} n(a)$
        \IF{an unused trace $r'_{b_j}\in\mathcal{R}_{b_j}$ exists}
            \STATE Form $\mathcal{S}^{\mathrm{probe}}_j$ by adding
            $(G_{\mathrm{norm}}(r'_{b_j}),b_j)$ to
            $\mathcal{S}^{\mathrm{ctrl}}$,
            preserving the order of the existing candidates
        \ELSE
            \STATE $\mathcal{S}^{\mathrm{probe}}_j
            \leftarrow\operatorname{Reverse}
            (\mathcal{S}^{\mathrm{ctrl}})$
        \ENDIF
        \STATE $p_j\leftarrow
        \mathcal{J}_j(q,\mathcal{S}^{\mathrm{probe}}_j)$
    \ENDFOR

    \hfill \textbf{Step 4: Final Aggregation}

    \STATE For each $a$, compute
    $H(a)=|\{j:c_j=p_j=a\}|$ and
    $S(a)=|\{j:p_j=a\}|$
    \IF{some $a$ satisfies $H(a)>0$ and $S(a)>J/2$}
        \STATE \textbf{return} $a$
    \ELSE
        \STATE \textbf{return} the strict-majority answer among
        $\{c_j\}_{j=1}^{J}$ if one exists; otherwise, the most-voted
        answer across $\{c_j\}_{j=1}^{J}$ and $\{p_j\}_{j=1}^{J}$
    \ENDIF
\end{algorithmic}
\end{algorithm}

\section{Experimental Settings}
\label{app:implementation}

Table~\ref{tab:hyperparameters} summarizes the main hyperparameters. For CoT generation, we follow the sampling settings recommended in the respective models' official model cards, where available, for temperature, top-$p$, and top-$k$.
All methods use the same CoT generation rule: generation stops after $k$ completed responses or $2k$ attempts per model, whichever occurs first. Truncated responses are discarded and resampled within the attempt
limit, and tokens from all attempts are included in the reported cost. We prompt the normalizer to target 85–115\% of the original length (Figure~\ref{fig:appendix-prompt-normalization}), while allowing a wider 80–120\% range when accepting normalized traces.

\begin{table}[t]
\centering
\small
\setlength{\tabcolsep}{8pt}
\renewcommand{\arraystretch}{1.05}

\caption{Hyperparameters used in the experiments.}
\label{tab:hyperparameters}

\begin{tabular}{@{}p{0.50\textwidth}cc@{}}
\toprule
\textbf{Parameter} & \multicolumn{2}{c}{\textbf{Value}} \\
\midrule

\rowcolor[HTML]{E8EEF8}
\multicolumn{3}{c}{\textbf{CoT generation}} \\
\midrule
\textbf{Model} & \textbf{Temperature} & \textbf{Top-$k$} \\
Phi-4-reasoning-plus          & 0.8 & 50 \\
Qwen3-8B                      & 0.6 & 20 \\
GLM-Z1-9B-0414                & 0.6 & 40 \\
NVIDIA-Nemotron-Nano-9B-v2    & 0.6 & Disabled \\
Top-$p$                       & \multicolumn{2}{c}{0.95} \\
Maximum output tokens         & \multicolumn{2}{c}{32,768} \\
Maximum generation attempts per model
& \multicolumn{2}{c}{Twice the target number of completed responses} \\

\midrule
\rowcolor[HTML]{E8EEF8}
\multicolumn{3}{c}{\textbf{Style normalization}} \\
\midrule
Rewriter                      & \multicolumn{2}{c}{Qwen3-8B} \\
Temperature                   & \multicolumn{2}{c}{0.6} \\
Top-$p$                       & \multicolumn{2}{c}{0.95} \\
Maximum output tokens         & \multicolumn{2}{c}{28,672} \\
Maximum retries (failed length or answer-preservation check)
& \multicolumn{2}{c}{2} \\
Accepted length ratio         & \multicolumn{2}{c}{$[0.8, 1.2]$} \\

\midrule
\rowcolor[HTML]{E8EEF8}
\multicolumn{3}{c}{\textbf{LLM decision}} \\
\midrule
Number of aggregating agents  & \multicolumn{2}{c}{3} \\
Temperature                   & \multicolumn{2}{c}{0.6} \\
Top-$p$                       & \multicolumn{2}{c}{0.95} \\
Maximum output tokens         & \multicolumn{2}{c}{32,768} \\
Maximum retries (invalid output)
& \multicolumn{2}{c}{2} \\
Maximum candidate answers per prompt
& \multicolumn{2}{c}{4} \\
Maximum trace length per candidate (characters)
& \multicolumn{2}{c}{6,000} \\

\midrule
\rowcolor[HTML]{E8EEF8}
\multicolumn{3}{c}{\textbf{Multi-agent baselines}} \\
\midrule
Temperature                   & \multicolumn{2}{c}{0.6} \\
Top-$p$                       & \multicolumn{2}{c}{0.95} \\
Maximum output tokens         & \multicolumn{2}{c}{32,768} \\
Maximum retries (truncated output)
& \multicolumn{2}{c}{4} \\
\bottomrule
\end{tabular}
\end{table}

\section{Extended Experimental Results}
\label{app:extended-results}

\subsection{Robustness to the Style Normalizer}
\label{app:normalizer-robustness}

We test whether the framework depends on Qwen3-8B as the style normalizer by replacing it with GLM-Z1-9B or Nemotron-Nano-9B while keeping the response pools and other settings fixed. Because the alternative normalizers were evaluated once, we use the first Qwen3-8B run as the matched reference rather than its three-run average. As shown in Table~\ref{tab:normalizer-robustness}, replacing Qwen3-8B with GLM-Z1-9B or Nemotron-Nano-9B changes average accuracy by $-2.4$ to $+1.3$ percentage points across the four model combinations. These results suggest that the framework can also be used with alternative style normalizers.

\begin{table*}[t]
\centering
\footnotesize
\setlength{\tabcolsep}{5pt}
\renewcommand{\arraystretch}{0.98}

\caption{\textbf{Final-answer accuracy with alternative style normalizers.}
Average is the unweighted mean across the three benchmarks, and $\Delta_{\mathrm{Qwen}}$ is its difference from the matched Qwen3-8B run. Alternative normalizers retain comparable performance across model combinations.}
\label{tab:normalizer-robustness}

\begin{tabular}{@{}llccccc@{}}
\toprule
Model combination & Style normalizer & OlymMATH & AIME & HMMT & Average & $\Delta_{\mathrm{Qwen}}$ (pp) \\
\midrule

\multirow{3}{*}{Qwen--Phi--GLM}
& Qwen3-8B     & 89.0 & 81.7 & 56.7 & 75.8 & -- \\
& GLM-Z1-9B               & 89.0 & 83.3 & 58.3 & 76.9 & $+1.1$ \\
& Nemotron-Nano-9B        & 93.0 & 81.7 & 56.7 & 77.1 & $+1.3$ \\
\midrule

\multirow{3}{*}{Qwen--Phi--Nemotron}
& Qwen3-8B     & 91.0 & 83.3 & 58.3 & 77.6 & -- \\
& GLM-Z1-9B               & 92.0 & 81.7 & 58.3 & 77.3 & $-0.3$ \\
& Nemotron-Nano-9B        & 90.0 & 80.0 & 63.3 & 77.8 & $+0.2$ \\
\midrule

\multirow{3}{*}{Qwen--GLM--Nemotron}
& Qwen3-8B     & 91.0 & 80.0 & 53.3 & 74.8 & -- \\
& GLM-Z1-9B               & 89.0 & 76.7 & 51.7 & 72.4 & $-2.4$ \\
& Nemotron-Nano-9B        & 88.0 & 80.0 & 51.7 & 73.2 & $-1.6$ \\
\midrule

\multirow{3}{*}{Phi--GLM--Nemotron}
& Qwen3-8B     & 89.0 & 85.0 & 61.7 & 78.6 & -- \\
& GLM-Z1-9B               & 87.0 & 83.3 & 60.0 & 76.8 & $-1.8$ \\
& Nemotron-Nano-9B        & 89.0 & 85.0 & 58.3 & 77.4 & $-1.2$ \\
\bottomrule
\end{tabular}
\end{table*}

\subsection{Tie-Breaking in SLM-MUX}
\label{app:slm-mux-tiebreaking}

SLM-MUX selects the modal answer of the model with the highest within-model consistency and originally resolves confidence ties using validation-set accuracy. Because our setting assumes no validation data, our main implementation instead applies label-free plurality voting over the pooled responses from tied models. To assess the effect of this substitution, we additionally evaluate a test-set Best-SC oracle that selects the tied model with the highest SC@$k$ accuracy on the corresponding test set. The two SLM-MUX variants use identical responses and inference cost. As shown in Table~\ref{tab:slm-mux-tiebreaking}, the oracle tie-breaker yields higher average accuracy than Pooled MV in all four model combinations, while Ours exceeds the oracle variant by 3.8--6.4 percentage points.

\begin{table*}[t]
\centering
\footnotesize
\setlength{\tabcolsep}{4pt}
\renewcommand{\arraystretch}{0.98}

\caption{\textbf{SLM-MUX under alternative confidence tie-breakers.}
The two SLM-MUX variants differ only in confidence tie-breaking and use identical responses and inference cost. Test-set Best-SC selects the tied model with the highest test-set SC@$k$ accuracy and serves as an oracle reference; Pooled MV is the label-free rule used in the main experiments. Ours is included for comparison, and bold marks the highest average.}
\label{tab:slm-mux-tiebreaking}

\begin{tabular}{@{}lllcccc@{}}
\toprule
Model combination & Method & Tie-breaker
& OlymMATH & AIME & HMMT & Average \\
\midrule

\multirow{3}{*}{Qwen--Phi--GLM}
& SLM-MUX & Test-set Best-SC (oracle) & 87.0 & 80.0 & 53.3 & 73.4 \\
\cmidrule(lr){2-7}
& SLM-MUX & Pooled MV (main)          & 86.0 & 78.3 & 48.3 & 70.9 \\
& \ourscell{Ours} & \ourscell{--}     & \ourscell{90.3} & \ourscell{83.3} & \ourscell{58.9} & \ourscell{\textbf{77.5}} \\
\midrule

\multirow{3}{*}{Qwen--Phi--Nemotron}
& SLM-MUX & Test-set Best-SC (oracle) & 88.0 & 78.3 & 55.0 & 73.8 \\
\cmidrule(lr){2-7}
& SLM-MUX & Pooled MV (main)          & 87.0 & 76.7 & 51.7 & 71.8 \\
& \ourscell{Ours} & \ourscell{--}     & \ourscell{90.7} & \ourscell{82.8} & \ourscell{59.4} & \ourscell{\textbf{77.6}} \\
\midrule

\multirow{3}{*}{Qwen--GLM--Nemotron}
& SLM-MUX & Test-set Best-SC (oracle) & 84.0 & 80.0 & 46.7 & 70.2 \\
\cmidrule(lr){2-7}
& SLM-MUX & Pooled MV (main)          & 85.0 & 76.7 & 46.7 & 69.4 \\
& \ourscell{Ours} & \ourscell{--}     & \ourscell{89.3} & \ourscell{78.9} & \ourscell{54.4} & \ourscell{\textbf{74.2}} \\
\midrule

\multirow{3}{*}{Phi--GLM--Nemotron}
& SLM-MUX & Test-set Best-SC (oracle) & 84.0 & 78.3 & 53.3 & 71.9 \\
\cmidrule(lr){2-7}
& SLM-MUX & Pooled MV (main)          & 87.0 & 78.3 & 48.3 & 71.2 \\
& \ourscell{Ours} & \ourscell{--}     & \ourscell{88.7} & \ourscell{85.0} & \ourscell{61.1} & \ourscell{\textbf{78.3}} \\

\bottomrule
\end{tabular}
\end{table*}

\subsection{Aggregation under Upstream Disagreement}
\label{app:upstream-disagreement}

We evaluate final-answer aggregation on instances not resolved by the upstream agreement gate in the independent-sampling setting with a target of $k=3$ completed responses per model. The evaluation subset is determined from these upstream responses before aggregation and is shared across all compared methods.

We reapply the budget-matching rule from Table~\ref{tab:main-results} using mean token costs on this subset: each baseline uses the lowest evaluated token budget at or above Ours. Ours uses $k=3$, while the selected baseline configurations are reported in Table~\ref{tab:upstream-disagreement}. LLM-based aggregation results are averaged over three runs, and rule-based methods are evaluated once. Single-Best-SC remains an oracle model-selection reference. Ours achieves the highest average accuracy in all four model combinations, with gains of 5.8--9.5 percentage points over the strongest non-oracle baseline.

\begin{table*}[t!]
\centering
\footnotesize
\setlength{\tabcolsep}{3.0pt}
\renewcommand{\arraystretch}{0.96}

\caption{\textbf{Final-answer accuracy and token usage under upstream disagreement.} Each baseline uses the nearest evaluated token budget at or above Ours on this subset. Average is the unweighted benchmark mean, and Tokens denotes input-and-output usage per problem. For LLM-based aggregation methods, values are means over three runs. Single-Best-SC is an oracle model-selection reference. Bold and underline mark the first- and second-ranked results, respectively.}
\label{tab:upstream-disagreement}

\begin{adjustbox}{width=\textwidth}
\begin{tabular}{clcccccc}
\toprule
Model combination & Method & Setting & OlymMATH $\uparrow$ & AIME $\uparrow$ & HMMT $\uparrow$ & Average $\uparrow$ & Tokens ($10^3$) $\downarrow$ \\
\midrule

\multirow{5}{*}{Qwen--Phi--GLM}
& Single-Best-SC~\citep{wang2022self} & $k=9$ & \underline{75.6} & \textbf{64.3} & 39.1 & \underline{59.7} & 380.3 \\
\cmidrule(lr){2-8}
& Combined MV & $k=4$ & 71.1 & \underline{57.1} & 34.8 & 54.3 & \underline{343.4} \\
& MAV~\citep{lifshitz2025multi} & 1 aspect & 61.5 & 41.7 & \underline{43.5} & 48.9 & 346.1 \\
& SLM-MUX~\citep{wang2026slm} & $k=4$ & 68.9 & 53.6 & 34.8 & 52.4 & \underline{343.4} \\
& \ourscell{Ours} & \ourscell{$k=3$} & \ourscell{\textbf{78.5}} & \ourscell{\textbf{64.3}} & \ourscell{\textbf{48.6}} & \ourscell{\textbf{63.8}} & \ourscell{\textbf{322.1}} \\
\midrule

\multirow{5}{*}{Qwen--Phi--Nemotron}
& Single-Best-SC & $k=9$ & 76.1 & \textbf{66.7} & 36.4 & \underline{59.7} & 383.8 \\
\cmidrule(lr){2-8}
& Combined MV & $k=4$ & \underline{78.3} & 55.6 & 40.9 & 58.2 & \underline{349.5} \\
& MAV & 1 aspect & 61.6 & 39.5 & \underline{41.7} & 47.6 & 350.3 \\
& SLM-MUX & $k=4$ & 71.7 & 51.9 & 36.4 & 53.3 & \underline{349.5} \\
& \ourscell{Ours} & \ourscell{$k=3$} & \ourscell{\textbf{79.7}} & \ourscell{\underline{65.4}} & \ourscell{\textbf{47.0}} & \ourscell{\textbf{64.0}} & \ourscell{\textbf{327.8}} \\
\midrule

\multirow{5}{*}{Qwen--GLM--Nemotron}
& Single-Best-SC & $k=9$ & \underline{79.2} & \textbf{55.6} & 31.7 & \underline{55.5} & \underline{230.4} \\
\cmidrule(lr){2-8}
& Combined MV & $k=4$ & 75.0 & 48.1 & 29.3 & 50.8 & 233.1 \\
& MAV & 1 aspect & 57.6 & 45.7 & \underline{34.1} & 45.8 & 237.1 \\
& SLM-MUX & $k=4$ & 70.8 & 48.1 & 29.3 & 49.4 & 233.1 \\
& \ourscell{Ours} & \ourscell{$k=3$} & \ourscell{\textbf{79.9}} & \ourscell{\underline{53.1}} & \ourscell{\textbf{40.7}} & \ourscell{\textbf{57.9}} & \ourscell{\textbf{229.8}} \\
\midrule

\multirow{5}{*}{Phi--GLM--Nemotron}
& Single-Best-SC & $k=9$ & 71.7 & \underline{67.7} & \underline{39.1} & \underline{59.5} & 396.4 \\
\cmidrule(lr){2-8}
& Combined MV & $k=4$ & \underline{76.1} & 61.3 & 34.8 & 57.4 & \underline{294.3} \\
& MAV & 1 aspect & 63.8 & 45.2 & \underline{39.1} & 49.4 & 307.8 \\
& SLM-MUX & $k=4$ & 73.9 & 58.1 & 34.8 & 55.6 & \underline{294.3} \\
& \ourscell{Ours} & \ourscell{$k=3$} & \ourscell{\textbf{77.5}} & \ourscell{\textbf{71.0}} & \ourscell{\textbf{51.4}} & \ourscell{\textbf{66.7}} & \ourscell{\textbf{279.9}} \\
\bottomrule
\end{tabular}
\end{adjustbox}
\end{table*}

\section{Additional Analyses}
\label{app:additional_analyses}

\subsection{Cross-Generator Convergence Across Representation Spaces}
\label{app:style_convergence}

\begin{figure*}[hb]
\centering
\includegraphics[width=0.80\textwidth]
{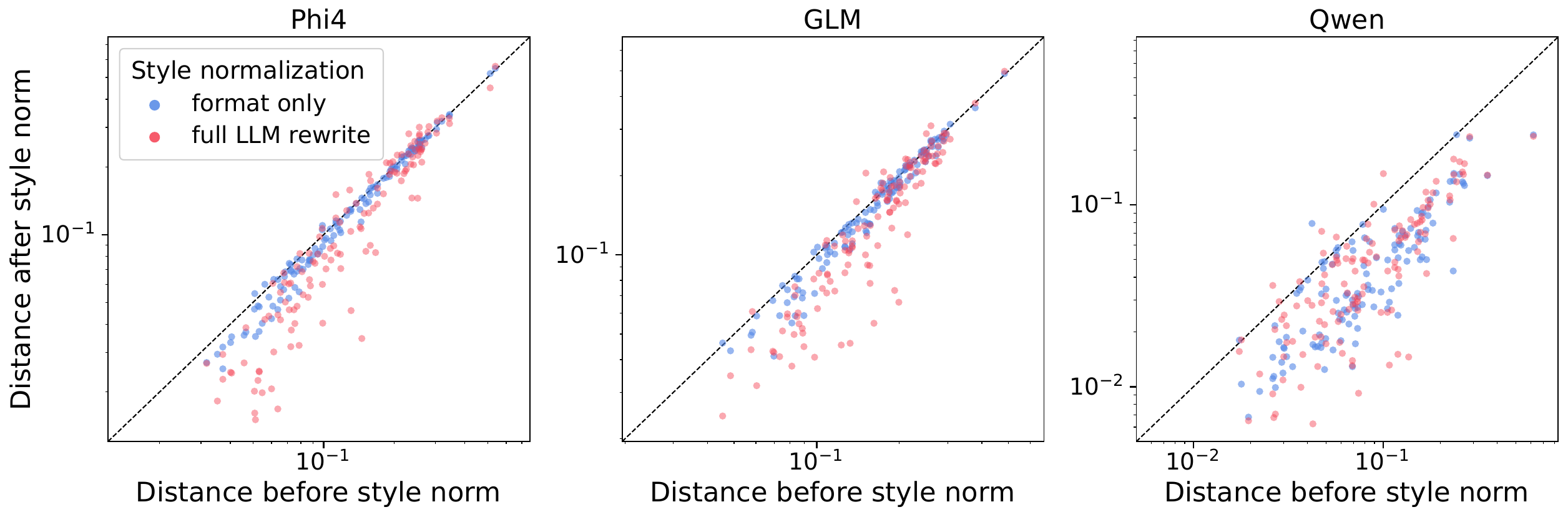}
\caption{\textbf{Cross-generator convergence across representation spaces.}
Panels show cosine distances in the Phi, GLM, and Qwen representation spaces between matched traces from different generators before and after formatting removal or full normalization. Traces are matched by problem and final answer. Points below the diagonal indicate reduced distance. Full normalization produces larger reductions than formatting removal, particularly in the Phi and GLM spaces.}
\label{fig:style_convergence_all}
\end{figure*}

Figure~\ref{fig:style_convergence_all} presents the full comparison across the Phi, GLM, and Qwen representation spaces. Formatting removal reduces mean representation distance by 1.7\% and 2.2\% in Phi and GLM, respectively, whereas full normalization reduces it by 8.2\% and 8.5\%.

\subsection{Source-Generator Attribution}
\label{app:source-attribution}

We use source attribution to diagnose which generator-dependent signals remain before and after normalization. For each feature channel, we train an L2-regularized multinomial logistic-regression classifier using reasoning traces. Evaluation uses five-fold cross-validation grouped by $(\text{dataset},\text{problem})$ to prevent the same problem from appearing in both training and test folds. Count features are length-normalized, and lexical vocabularies are fit on the original traces and reused after normalization. Table~\ref{tab:generator-signals} reports attribution accuracy before and after normalization across the six feature channels.

Normalization reduces attribution accuracy most strongly for formatting features (94.1\% to 72.2\%), with smaller reductions for rhetorical markers (88.8\% to 79.7\%) and lexical features. These decreases suggest that the corresponding feature sets become less informative about the source model, although substantial source information remains after normalization. They measure source distinguishability within feature groups, rather than each group's contribution to representation-level convergence.

\begin{table}[t]
    \centering
    \small
    \caption{\textbf{Source-generator attribution before and after style normalization.}
    Values report three-way classification accuracy under grouped five-fold
    cross-validation; $\Delta$ uses unrounded values. Normalization attenuates
    formatting cues most strongly, while substantial source signal remains.}
    \label{tab:generator-signals}
    \begin{tabular}{lccc}
        \toprule
        Feature channel & Original & Normalized & $\Delta$ \\
        \midrule
        Formatting            & 94.1  & 72.2 & $-21.9$ \\
        Rhetorical markers    & 88.8  & 79.7 & $-9.1$ \\
        Character 3-grams     & 99.2  & 91.9 & $-7.3$ \\
        Bag of words          & 94.6  & 88.5 & $-6.1$ \\
        Function words        & 86.6  & 80.9 & $-5.7$ \\
        Judge representation & 100.0 & 97.2 & $-2.8$ \\
        \bottomrule
    \end{tabular}
\end{table}

\subsection{Additional Analysis of Final Aggregation}
\label{app:final_aggregation}

\begin{table}[h]
\centering
\small
\setlength{\tabcolsep}{10pt}
\renewcommand{\arraystretch}{1.05}
\caption{\textbf{Answer accuracy by hold and support.}
Conditional accuracy (\%) by $H(a)$ and $S(a)$.
$J$ denotes the number of agents.}
\label{tab:hold_support}
\begin{tabular}{@{}lcc@{}}
\toprule
& $H(a)=0$ & $H(a)>0$ \\
\midrule
$S(a)\le J/2$ & 13.3 & 23.5 \\
$S(a)>J/2$    & 18.1 & 49.1 \\
\bottomrule
\end{tabular}
\end{table}

On instances where the controlled decisions are not unanimous and adverse-frequency injection is applied, we analyze candidate-answer accuracy conditioned on hold count $H(a)$ and second-pass support count $S(a)$. Here, $H(a)$ counts judges selecting answer $a$ in both the controlled and probe passes, while $S(a)$ counts probe judgments selecting $a$. As shown in Table~\ref{tab:hold_support}, answers with both a hold and majority probe support have the highest conditional accuracy among the four groups: 49.1\%, compared with 18.1\% for majority support without a hold and 23.5\% for a hold without majority support.

\clearpage
\subsection{Examples of Style Normalization}
\label{app:normalization-examples}

Tables~\ref{tab:normalization-example-correct}
and~\ref{tab:normalization-example-competing} present the complete original
and normalized traces supporting the
correct answer \(6\) and the competing answer \(1\), respectively, from the example illustrated in
Figure~\ref{fig:main-modules}. In both examples, normalization changes wording and presentation
while preserving the main reasoning steps, calculations, and final answers,
including the errors in the competing trace. The normalization prompt
contained the original trace and rewriting instructions; the problem
statement was not included as a separate input. Without style normalization, agents judge the original traces (shown in red and blue); with normalization, they judge versions with more consistent presentation (shown in gray).

\begingroup
\normalsize
\definecolor{NormExampleBlue}{HTML}{DBE6FA}
\definecolor{NormExampleRed}{HTML}{FED8DC}
\definecolor{NormExampleGray}{HTML}{F2F2F2}
\setlength{\LTleft}{0pt}
\setlength{\LTright}{0pt plus 1fill}
\setlength{\LTcapwidth}{\textwidth}

\begin{longtable}{@{\hspace{4pt}}>{\columncolor{NormExampleRed}[4pt][4pt]\raggedright\arraybackslash}p{\dimexpr(\textwidth-1em-16pt)/2\relax}@{\hspace{\dimexpr1em+8pt\relax}}>{\columncolor{NormExampleGray}[4pt][4pt]\raggedright\arraybackslash}p{\dimexpr(\textwidth-1em-16pt)/2\relax}@{\hspace{4pt}}}
\caption{\textbf{Complete original and normalized traces for the correct answer \(\boxed{6}\).}}
\label{tab:normalization-example-correct}\\
\textbf{Original trace} & \textbf{Style-normalized trace} \\
\endfirsthead
\multicolumn{2}{@{}l@{}}{\tablename~\thetable\ (continued)}\\
\textbf{Original trace} & \textbf{Style-normalized trace} \\
\endhead
\hline
\multicolumn{2}{r@{}}{Continued on next page}\\
\endfoot
\hline
\endlastfoot
To determine the number of ways to divide a \(20 \times 24\) rectangle into
\(4 \times 5\) rectangles, we first verify the feasibility of such a tiling:

\smallskip
\textbf{Feasibility Check}

\(\bullet\) \textbf{Area of the large rectangle:}
\[
20 \times 24 = 480
\]

\(\bullet\) \textbf{Area of each tile:}
\[
4 \times 5 = 20
\]

\(\bullet\) \textbf{Number of tiles needed:}
\[
\frac{480}{20} = 24
\]

Since the area is divisible and the tile dimensions are smaller than the
rectangle, a tiling is possible. However, we must also ensure that the tiling
can be done with the given dimensions, accounting for the fact that the tile
can be rotated (i.e., placed as \(4 \times 5\) or \(5 \times 4\)).
&
To determine the number of ways to divide a \(20 \times 24\) rectangle into
\(4 \times 5\) rectangles, we first verify the feasibility of such a tiling.
The area of the large rectangle is \(20 \times 24 = 480\). Each tile has an
area of \(4 \times 5 = 20\), so the number of tiles required is
\(480 \div 20 = 24\). Since the area is divisible and the tile dimensions
are smaller than the rectangle, a tiling is possible. However, we must also
ensure that the tiling can be done with the given dimensions, accounting for
the fact that the tile can be rotated.
\\[0.6em]

\smallskip
\textbf{Key Insight: Tiling via Segmentations}

To tile the entire rectangle, we can divide the \textbf{width} (24 units)
into segments of either 4 or 5 units. Each such segment can then be filled
with tiles of the appropriate orientation:

\(\bullet\) A segment of \textbf{4 units wide} can be tiled with
\textbf{vertical tiles} (5 units tall, 4 units wide).

\(\bullet\) A segment of \textbf{5 units wide} can be tiled with
\textbf{horizontal tiles} (4 units tall, 5 units wide).

Since the height of the rectangle is 20, which is divisible by both 4 and 5,
we can tile each such segment with tiles of the correct orientation.
Therefore, the problem reduces to finding the number of
\textbf{compositions} of 24 into parts of 4 and 5, where the
\textbf{order of the segments matters} (since different arrangements
correspond to different tilings).
&
To tile the entire rectangle, we can divide the width (24 units) into
segments of either 4 or 5 units. Each such segment can then be filled with
tiles of the appropriate orientation. A segment of 4 units wide can be tiled
with vertical tiles (5 units tall, 4 units wide). A segment of 5 units wide
can be tiled with horizontal tiles (4 units tall, 5 units wide). Since the
height of the rectangle is 20, which is divisible by both 4 and 5, we can
tile each such segment with tiles of the correct orientation. Therefore, the
problem reduces to finding the number of compositions of 24 into parts of
4 and 5, where the order of the segments matters.
\\[0.6em]

\smallskip
\textbf{Counting the Number of Valid Compositions}

We define a function \(C(n)\) as the number of compositions of \(n\) into
parts of 4 and 5. This satisfies the recurrence:
\[
C(n) = C(n-4) + C(n-5)
\]
with base cases:

\(\bullet\) \(C(0)=1\) (the empty composition),

\(\bullet\) \(C(n)=0\) for \(n<0\).
&
We define a function \(C(n)\) as the number of compositions of \(n\) into
parts of 4 and 5. This satisfies the recurrence:
\[
C(n) = C(n-4) + C(n-5)
\]
with base cases:

\(\bullet\) \(C(0)=1\) (the empty composition),

\(\bullet\) \(C(n)=0\) for \(n<0\).
\\[0.6em]

We compute \(C(24)\) using this recurrence:\par
\(\begin{aligned}[t]
C(0) &= 1 \\
C(1) &= 0 \\
C(2) &= 0 \\
C(3) &= 0 \\
C(4) &= C(0) = 1
\end{aligned}\)
&
We compute \(C(24)\) using this recurrence:\par
\(\begin{aligned}[t]
C(0) &= 1 \\
C(1) &= 0 \\
C(2) &= 0 \\
C(3) &= 0 \\
C(4) &= C(0) = 1
\end{aligned}\)
\\[0.15em]

\(\begin{aligned}[t]
C(5) &= C(1)+C(0) = 0+1 = 1 \\
C(6) &= C(2)+C(1) = 0+0 = 0 \\
C(7) &= C(3)+C(2) = 0+0 = 0 \\
C(8) &= C(4)+C(3) = 1+0 = 1 \\
C(9) &= C(5)+C(4) = 1+1 = 2
\end{aligned}\)
&
\(\begin{aligned}[t]
C(5) &= C(1)+C(0) = 0+1 = 1 \\
C(6) &= C(2)+C(1) = 0+0 = 0 \\
C(7) &= C(3)+C(2) = 0+0 = 0 \\
C(8) &= C(4)+C(3) = 1+0 = 1 \\
C(9) &= C(5)+C(4) = 1+1 = 2
\end{aligned}\)
\\[0.15em]

\(\begin{aligned}[t]
C(10) &= C(6)+C(5) = 0+1 = 1 \\
C(11) &= C(7)+C(6) = 0+0 = 0 \\
C(12) &= C(8)+C(7) = 1+0 = 1 \\
C(13) &= C(9)+C(8) = 2+1 = 3 \\
C(14) &= C(10)+C(9) = 1+2 = 3
\end{aligned}\)
&
\(\begin{aligned}[t]
C(10) &= C(6)+C(5) = 0+1 = 1 \\
C(11) &= C(7)+C(6) = 0+0 = 0 \\
C(12) &= C(8)+C(7) = 1+0 = 1 \\
C(13) &= C(9)+C(8) = 2+1 = 3 \\
C(14) &= C(10)+C(9) = 1+2 = 3
\end{aligned}\)
\\[0.15em]

\(\begin{aligned}[t]
C(15) &= C(11)+C(10) = 0+1 = 1 \\
C(16) &= C(12)+C(11) = 1+0 = 1 \\
C(17) &= C(13)+C(12) = 3+1 = 4 \\
C(18) &= C(14)+C(13) = 3+3 = 6 \\
C(19) &= C(15)+C(14) = 1+3 = 4
\end{aligned}\)
&
\(\begin{aligned}[t]
C(15) &= C(11)+C(10) = 0+1 = 1 \\
C(16) &= C(12)+C(11) = 1+0 = 1 \\
C(17) &= C(13)+C(12) = 3+1 = 4 \\
C(18) &= C(14)+C(13) = 3+3 = 6 \\
C(19) &= C(15)+C(14) = 1+3 = 4
\end{aligned}\)
\\[0.15em]

\(\begin{aligned}[t]
C(20) &= C(16)+C(15) = 1+1 = 2 \\
C(21) &= C(17)+C(16) = 4+1 = 5 \\
C(22) &= C(18)+C(17) = 6+4 = 10 \\
C(23) &= C(19)+C(18) = 4+6 = 10 \\
C(24) &= C(20)+C(19) = 2+4 = 6
\end{aligned}\)
&
\(\begin{aligned}[t]
C(20) &= C(16)+C(15) = 1+1 = 2 \\
C(21) &= C(17)+C(16) = 4+1 = 5 \\
C(22) &= C(18)+C(17) = 6+4 = 10 \\
C(23) &= C(19)+C(18) = 4+6 = 10 \\
C(24) &= C(20)+C(19) = 2+4 = 6
\end{aligned}\)
\\[0.15em]

Thus, the number of valid tilings is:
\[
\boxed{6}
\]
&
Thus, the number of valid tilings is:
\[
\boxed{6}
\]
\\[0.6em]

\smallskip
\textbf{Conclusion}

Each valid composition of 24 into parts of 4 and 5 corresponds to a unique
way of dividing the rectangle into segments that can be tiled with
\(4 \times 5\) tiles in appropriate orientations. Since the problem considers
rotations and reflections as distinct, each composition yields a distinct
tiling.
&
Each valid composition of 24 into parts of 4 and 5 corresponds to a unique
way of dividing the rectangle into segments that can be tiled with
\(4 \times 5\) tiles in appropriate orientations. Since the problem considers
rotations and reflections as distinct, each composition yields a distinct
tiling.
\\*[0.6em]

\smallskip
\textbf{Final Answer}
\[
\boxed{6}
\]
&
\[
\boxed{6}
\]
\\[0.6em]
\end{longtable}

\clearpage

\begin{longtable}{@{\hspace{4pt}}>{\columncolor{NormExampleBlue}[4pt][4pt]\raggedright\arraybackslash}p{\dimexpr(\textwidth-1em-16pt)/2\relax}@{\hspace{\dimexpr1em+8pt\relax}}>{\columncolor{NormExampleGray}[4pt][4pt]\raggedright\arraybackslash}p{\dimexpr(\textwidth-1em-16pt)/2\relax}@{\hspace{4pt}}}
\caption{\textbf{Complete original and normalized traces for the competing answer \(\boxed{1}\).}}
\label{tab:normalization-example-competing}\\
\textbf{Original trace} & \textbf{Style-normalized trace} \\
\endfirsthead
\multicolumn{2}{@{}l@{}}{\tablename~\thetable\ (continued)}\\
\textbf{Original trace} & \textbf{Style-normalized trace} \\
\endhead
\hline
\multicolumn{2}{r@{}}{Continued on next page}\\
\endfoot
\hline
\endlastfoot
We will show that no ``exotic'' tiling exists--in fact, there is only one
way to cover a \(20\times24\) rectangle by \(4\times5\) rectangles (when
rotations and reflections count as different, so that even a ``mirrored''
copy would be considered distinct--but here there is no choice at all).
&
We aim to demonstrate that no unconventional tiling is possible--in fact,
there is only one way to cover a \(20\times24\) rectangle using
\(4\times5\) rectangles (when rotations and reflections count as distinct,
so that even a mirrored copy would be considered different--but here there
is no choice at all).
\\[0.6em]

Let us explain.

A \(4\times5\) rectangle has area 20. Since
\[
\begin{gathered}
20\times24=480
\\
\qquad\text{and}\qquad
480\div20=24,
\end{gathered}
\]
any tiling must use exactly 24 tiles.
&
First, let us explain. A \(4\times5\) rectangle has an area of 20. Given that
\(20\times24=480\) and 480 divided by 20 equals 24, any tiling requires
exactly 24 tiles.
\\[0.6em]

Moreover, since all tiles are congruent (up to rotation) and their sides must
be parallel to the sides of the large rectangle, one may show (by a standard
``grid-forcing'' argument) that if a rectangle is tiled by congruent
(non-square) rectangles then the tiling ``extends'' to a complete set of
horizontal and vertical lines. In other words, every tiling by congruent
non-square rectangles is necessarily a grid tiling.
&
Moreover, since all tiles are congruent (up to rotation) and their sides must
align with the sides of the large rectangle, it can be shown (via a standard
grid-forcing argument) that tiling a rectangle with congruent non-square
rectangles necessitates a grid-like structure. In other words, every tiling
by congruent non-square rectangles is necessarily a grid tiling.
\\[0.6em]

Now, note that a \(4\times5\) rectangle has two possible orientations. In one
orientation the longer side (of length 5) is vertical and the shorter side
(of length 4) is horizontal; in the other the roles are reversed. But look
at the dimensions of the big rectangle:

\(\bullet\) The vertical side is 20 and the horizontal side is 24.
&
Now, note that a \(4\times5\) rectangle can be oriented in two ways: one with
the longer side (length 5) vertical and the shorter side (length 4)
horizontal, and the other with these dimensions reversed. But consider the
dimensions of the large rectangle:

\(\bullet\) The vertical side is 20 and the horizontal side is 24.
\\[0.6em]

If we try to fit a \(4\times5\) rectangle with the 5-side vertical, then its
horizontal side is 4. Since 24 is divisible by 4 (\(24\div4=6\)) but 20 is
not divisible by 4 (\(20\div4=5\), and then \(5\times4=20\) would force the
other dimension to be 5, not 20), the only possibility is to have every tile
placed in the orientation in which the 5-side is vertical and the 4-side is
horizontal. In that case the vertical cell has height 5 and the horizontal
cell has width 4. Then the tiling is given by
&
If we attempt to fit a \(4\times5\) rectangle with the 5-side vertical, its
horizontal side would be 4. Since 24 is divisible by 4 (\(24\div4=6\)) but
20 is not divisible by 4 (\(20\div4=5\), and \(5\times4=20\) would force
the other dimension to be 5, not 20), the only viable orientation is one in
which the 5-side is vertical and the 4-side is horizontal. In this case, the
vertical cell has height 5 and the horizontal cell has width 4. Then the
tiling is achieved by
\\[0.6em]

\(\bullet\) dividing the 20-side into \(20\div5=4\) equal segments, and

\(\bullet\) dividing the 24-side into \(24\div4=6\) equal segments.

This produces a \(4\times6\) grid of \(5\times4\) rectangles.
&
\(\bullet\) dividing the 20-side into \(20\div5=4\) equal segments, and

\(\bullet\) dividing the 24-side into \(24\div4=6\) equal segments.

This produces a \(4\times6\) grid of \(5\times4\) rectangles.
\\[0.6em]

One may check that no ``staggered'' or ``non-grid'' tiling is possible (in
fact, one may prove that any tiling of a rectangle by congruent non-square
rectangles must be a grid tiling). Also, if one were to try to use the other
orientation--with the 4-side vertical--then the horizontal side would be 5,
but 24 is not divisible by 5.
&
One may verify that no staggered or non-grid tiling is feasible (in fact, one
may prove that any tiling of a rectangle by congruent non-square rectangles
must be a grid tiling). Additionally, if one were to try the other
orientation--with the 4-side vertical--the horizontal side would be 5, but
24 is not divisible by 5.
\\[0.6em]

Thus the tiling is unique. Since rotations and reflections are considered
distinct in the statement, even if one were to flip the entire tiling, one
would still have the same grid; there is no alternative.
&
Thus the tiling is unique. Since rotations and reflections are considered
distinct in the statement, even if one were to flip the entire tiling, the
grid remains unchanged; there is no alternative.
\\[0.6em]

Therefore, the number of ways to divide a \(20\times24\) rectangle into
\(4\times5\) rectangles is
\[
\boxed{1}.
\]
&
Therefore, the number of ways to divide a \(20\times24\) rectangle into
\(4\times5\) rectangles is
\[
\boxed{1}.
\]
\\[0.6em]
\end{longtable}

\endgroup

\clearpage
\section{Prompts}
\label{sec:app_prompts}

We provide the prompts used for style normalization and final-answer decision. Figure~\ref{fig:appendix-prompt-normalization} shows the style-normalization prompt, and Figure~\ref{fig:appendix-prompt-judgment-frequency} illustrates the decision prompt before and after adverse-frequency bias injection. The bias injection stage uses the same decision instruction as the bias-controlled stage; only the candidate presentation is changed.

\begin{figure*}[h]
\centering

\begin{tcolorbox}[
    colback=col!5,
    colframe=col,
    boxrule=0.8pt,
    arc=2pt,
    left=4pt,right=4pt,top=4pt,bottom=4pt,
    width=0.97\textwidth
]
\begin{lstlisting}[style=promptstyle]
Target style: a single uniform plain-prose voice.
  - No markdown headers (no '#', no '##').
  - No bullet lists; write in continuous sentences.
  - LaTeX only for inline equations and any \boxed{...}; do NOT decorate prose.
  - Short paragraphs; transitions like "First,", "Next,", "Therefore,".
  - Preserve any \boxed{...} exactly.

LENGTH (hard): the passage is {n} chars; keep your rewrite within {0.85n}-{1.15n} chars.

=== Rewrite the following passage in the target style ===
Rewrite ONLY the surface wording, phrasing, and formatting. 
Keep ALL content exactly as it is: 
every statement, number, equation, and \boxed{...} stays unchanged in meaning and value.
Do NOT add anything that is not already there -- no commentary, no confirmations or checks
("this is correct", "this checks out", "consistent with", "indeed", ...), no corrections, no flags.
Do NOT remove or reorder content. If something looks off, leave it exactly as written.
You are only changing how it reads, not what it says.

Preserve any final \boxed{...} exactly (character-for-character). Output ONLY the rewritten passage, nothing else.

## Passage ({n} chars)
{trace}

## Your rewrite (target: {0.85n}-{1.15n} chars)
\end{lstlisting}
\end{tcolorbox}
\caption{\textbf{Style-normalization prompt.}
The prompt instructs the rewriter to change only wording and formatting while preserving reasoning content and final answers; a length constraint limits deviation from the original trace.}
\label{fig:appendix-prompt-normalization}
\end{figure*}

\begin{figure*}[t]
\centering

\begin{subfigure}[t]{0.97\textwidth}
\centering
\begin{tcolorbox}[
    colback=col!5,
    colframe=col,
    boxrule=0.8pt,
    arc=2pt,
    left=4pt,right=4pt,top=4pt,bottom=4pt,
    width=\textwidth
]
\begin{lstlisting}[style=promptstyle]
You are given several candidate solutions (anonymized) to the same problem. Decide which one's final answer is correct. If you believe none of them is correct, answer none.

Problem:
Eight circles of radius $34$ are sequentially tangent, and two of the circles are tangent to $AB$ and $BC$ of triangle $ABC$, respectively. $2024$ circles of radius $1$ can be arranged in the same manner. The inradius of triangle $ABC$ can be expressed as $\frac{m}{n}$, where $m$ and $n$ are relatively prime positive integers. Find $m+n$.

--- Solution 0 (answer: 2297) ---
[FULL NORMALIZED REASONING TRACE FOR 2297]

--- Solution 1 (answer: 254) ---
[FULL NORMALIZED REASONING TRACE FOR 254]

--- Solution 2 (answer: 197) ---
[FULL NORMALIZED REASONING TRACE #1 FOR 197]

--- Solution 3 (answer: 8503) ---
[FULL NORMALIZED REASONING TRACE FOR 8503]

Reason step by step about which final answer is correct, then put the chosen final answer within \boxed{}, or \boxed{none} if none of the candidates' answers is correct.
\end{lstlisting}
\end{tcolorbox}
\caption{Prompt for the initial bias-controlled decision.}
\label{fig:appendix-prompt-judgment-controlled}
\end{subfigure}

\vspace{6pt}

\begin{subfigure}[t]{0.97\textwidth}
\centering
\begin{tcolorbox}[
    colback=col!5,
    colframe=col,
    boxrule=0.8pt,
    arc=2pt,
    left=4pt,right=4pt,top=4pt,bottom=4pt,
    width=\textwidth
]
\begin{lstlisting}[style=promptstyle]
You are given several candidate solutions (anonymized) to the same problem. Decide which one's final answer is correct. If you believe none of them is correct, answer none.

Problem:
Eight circles of radius $34$ are sequentially tangent, and two of the circles are tangent to $AB$ and $BC$ of triangle $ABC$, respectively. $2024$ circles of radius $1$ can be arranged in the same manner. The inradius of triangle $ABC$ can be expressed as $\frac{m}{n}$, where $m$ and $n$ are relatively prime positive integers. Find $m+n$.

--- Solution 0 (answer: 2297) ---
[FULL NORMALIZED REASONING TRACE FOR 2297]

--- Solution 1 (answer: 254) ---
[FULL NORMALIZED REASONING TRACE FOR 254]

(@{\color{promptblue}\bfseries\ttfamily --- Solution 2 (answer: 197) ---}@)
(@{\color{promptblue}\ttfamily [FULL NORMALIZED REASONING TRACE \#1 FOR 197]}@)

--- Solution 3 (answer: 8503) ---
[FULL NORMALIZED REASONING TRACE FOR 8503]

(@{\color{promptblue}\bfseries\ttfamily --- Solution 4 (answer: 197) ---}@)
(@{\color{promptblue}\ttfamily [FULL NORMALIZED REASONING TRACE \#2 FOR 197]}@)

Reason step by step about which final answer is correct, then put the chosen final answer within \boxed{}, or \boxed{none} if none of the candidates' answers is correct.
\end{lstlisting}
\end{tcolorbox}
\caption{Prompt for the second-pass decision under adverse-frequency bias injection.}
\label{fig:appendix-prompt-judgment-adverse}
\end{subfigure}

\caption{\textbf{Decision prompts before and after adverse-frequency bias injection.}
(a) Bias-controlled presentation with one trace per answer hypothesis.
(b) After the controlled decision selects an answer other than $197$, an additional existing trace makes $197$ the higher-frequency competitor without changing the problem, answer hypotheses, or decision instruction.}
\label{fig:appendix-prompt-judgment-frequency}
\end{figure*}

\end{document}